\documentclass[runningheads]{llncs}
\usepackage{graphicx}
\usepackage{tikz}
\usepackage{comment}
\usepackage{amsmath,amssymb} % define this before the line numbering.
\usepackage{color}

\usepackage{booktabs}
\usepackage{epsfig}
\usepackage{multicol}
\usepackage{multirow}
\usepackage{arydshln}
\usepackage{pifont}
\usepackage{float}
\usepackage{color, colortbl}
\usepackage{xcolor}
\usepackage{babel}
\usepackage{capt-of}

\usepackage{xspace}

\usepackage[accsupp]{axessibility}  % Improves PDF readability for those with disabilities.

\usepackage[pagebackref,breaklinks,colorlinks]{hyperref}

\definecolor{LightYellow}{rgb}{0.99,0.94,0.7}
\definecolor{LightCyan}{rgb}{0.88,1,1}

\newcommand\blfootnote[1]{%
  \begingroup
  \renewcommand\thefootnote{}\footnote{#1}%
  \addtocounter{footnote}{-1}%
  \endgroup
}

\makeatletter
\DeclareRobustCommand\onedot{\futurelet\@let@token\@onedot}
\def\@onedot{\ifx\@let@token.\else.\null\fi\xspace}
\def\eg{\emph{e.g}\onedot} 
\def\ie{\emph{i.e}\onedot}

\makeatother

\begin{document}
\pagestyle{headings}
\mainmatter
\def\ECCVSubNumber{4300}  % Insert your submission number here

\title{Constrained Mean Shift Using Distant Yet \\ Related Neighbors for Representation Learning}

%******************

\titlerunning{Constrained Mean Shift Using Distant Yet Related Neighbors}
\author{K L Navaneet\inst{1} $^{*}$ \index{Navaneet, K L} \and Soroush Abbasi Koohpayegani\inst{1} $^{*}$ \index{Abbasi Koohpayegani, Soroush} \and Ajinkya Tejankar\inst{1} $^{*}$\and Kossar Pourahmadi\inst{1}\and Akshayvarun Subramanya\inst{2} \and Hamed Pirsiavash\inst{1}
 }
\authorrunning{Navaneet, Abbasi Koohpayegani, Tejankar et al.}

 \institute{
University of California, Davis \and University of Maryland, Baltimore County 
}
%******************
\maketitle

%%%%%%%%% ABSTRACT
\begin{abstract}
We are interested in representation learning in self-supervised, supervised, and semi-supervised settings. Some recent self-supervised learning methods like mean-shift (MSF) cluster images by pulling the embedding of a query image to be closer to its nearest neighbors (NNs). Since most NNs are close to the query by design, the averaging may not affect the embedding of the query much. On the other hand, far away NNs may not be semantically related to the query. We generalize the mean-shift idea by constraining the search space of NNs using another source of knowledge so that NNs are far from the query while still being semantically related. We show that our method (1) outperforms MSF in SSL setting when the constraint utilizes a different augmentation of an image from the previous epoch, and (2) outperforms PAWS in semi-supervised setting with less training resources when the constraint ensures that the NNs have the same pseudo-label as the query. Our code is available here: \textcolor{magenta}{\href{https://github.com/UCDvision/CMSF}{https://github.com/UCDvision/CMSF}}

\end{abstract}

%%%%%%%%% BODY TEXT
\section{Introduction}

Recently, we have seen great progress in self-supervised learning (SSL) methods that learn rich representations from unlabeled data. Such methods are important since they do not rely on manual annotation of data, which can be costly, biased, or ambiguous. Hence, SSL representations may perform better than supervised ones in transferring to downstream visual recognition tasks.

\blfootnote{* equal contribution}

\begin{figure}
    \centering
    \begin{minipage}{0.5\textwidth}
        \centering
        \includegraphics[width=1.0\linewidth]{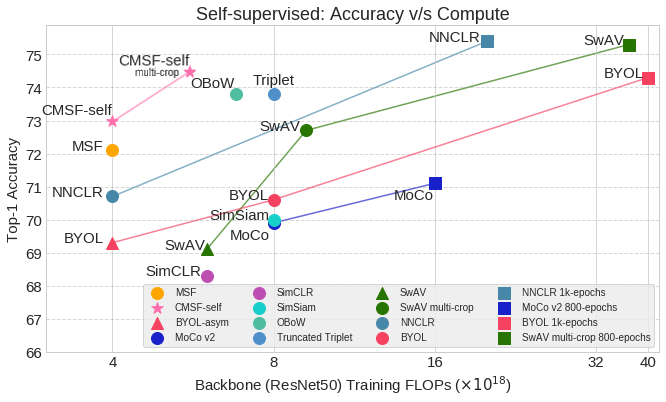}
    \end{minipage}%
    \begin{minipage}{0.5\textwidth}
        \centering
        \includegraphics[width=1.0\linewidth]{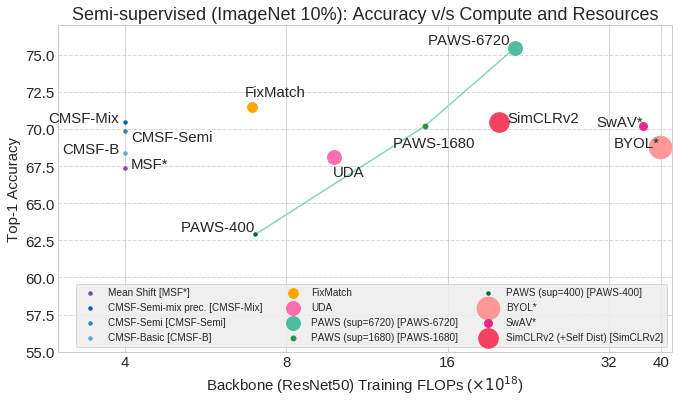}
    \end{minipage}
    \caption{\textbf{Accuracy vs. training compute on ImageNet with ResNet50:} We report the total training FLOPs for forward and backward passes through the CNN backbone. \textbf{(Left) Self-supervised: }All methods are for 200 epochs. $\textrm{CMSF}_{\textrm{self}}$ achieves competitive accuracy with considerably lower compute. \textbf{(Right) Semi-supervised:} Circle radius is proportional to the number of GPUs/TPUs used. The results are on ImageNet with $10\%$ labels. In addition to being compute efficient, CMSF is trained with an order of magnitude lower resources, making it more practical and accessible. * methods use self-supervised pre-training and finetuning on the labeled set.} 
    \label{fig:compute_semi_sup}
\end{figure}

\begin{figure*}[th]
\centering
\includegraphics[width=0.95\linewidth]{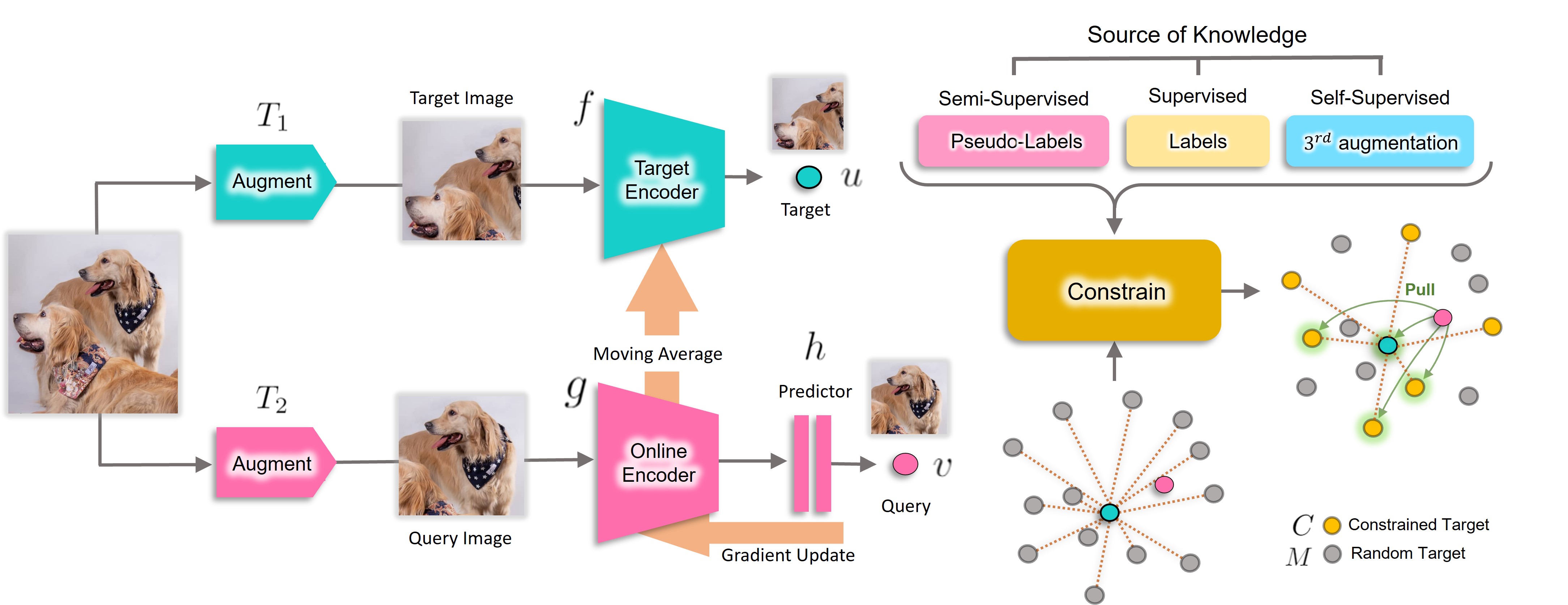}
\caption{{\bf Our method (CMSF):} We augment an image twice and pass them through online and target encoders followed by $\ell_2$ normalization to get $u$ and $v$. Mean-shift~\cite{koohpayegani2021mean} encourages $v$ to be close to both $u$ and its nearest neighbors (NN). To make NNs diverse, we constrain the NN search space based on additional knowledge in the form of NNs of the previous augmentation in self-supervised setting or the labels or pseudo-labels in semi or fully supervised settings. These constraints encourages the query to be pulled towards semantically related NNs that are farther away from the target embedding. See Fig \ref{fig:2q_teaser} for constructing the constrained set.} 
\label{fig:teaser}

\end{figure*}

Most recent SSL methods, \eg, MoCo \cite{he2020momentum} and BYOL \cite{grill2020bootstrap}, pull the embedding of a query image to be closer to its own augmentation compared to some other random images. Follow-up works have focused on improving the positive pairs through generating better augmentations~\cite{tian2020makes,reed2021selfaugment,lee2020mix} and the negative set by increasing the set size~\cite{he2020momentum} or mining effective samples~\cite{kalantidis2020hard,huynh2020boosting,wang2021solving}, but have largely ignored possibility of utilizing additional positive images. More recently, ~\cite{koohpayegani2021mean,dwibedi2021little,azabou2021mine} expand the positive set using nearest neighbors.
Inspired by classic mean-shift algorithm, MSF \cite{koohpayegani2021mean} generalizes BYOL to group similar images together. MSF pulls a query image to be close to not only its augmentation, but also the top-$k$ nearest neighbors (NNs) of its augmentation. 

We argue that the top-$k$ neighbors are close to the query image by construction, and thus may not provide a strong supervision signal. We are interested in choosing far away (non-top) neighbors that are still semantically related to the query image. This cannot be trivially achieved by increasing the number of NNs since the {\em purity} of retrieved neighbors decreases with increasing $k$ (See Fig.~\ref{fig:visualize_nns} and Fig.~\ref{fig:purity}). Purity is defined as the percentage of the NNs belonging to the same category as the query image. 

We generalize MSF~\cite{koohpayegani2021mean} method by simply limiting the NN search to a smaller subset that we believe is reasonably far from the query but still semantically related to it. We define this constraint to be (1) the nearest neighbors of another augmentation of the query in SSL setting and (2) images sharing the same label or pseudo-label as the query in supervised and semi-supervised settings. While we aim to obtain distant samples of the same category, note that we group only a few neighbors ($k$ in our method) from the constrained subset instead of grouping the whole subset together. This is in contrast to cross-entropy supervised learning, where we pull all images of a category to form a cluster or be on the same side of a hyper-plane. Our method can benefit from this relaxation by preserving the latent structure of the categories and also being robust to noisy labels.

Our experiments show that the method outperforms the various baselines in all three settings with same or less amount of computation in training (refer Fig.~\ref{fig:compute_semi_sup}). It outperforms MSF \cite{koohpayegani2021mean} in SSL, cross-entropy in supervised (with clean or noisy labels), and PAWS \cite{assran2021semi} in semi-supervised settings. Our main novelty is in developing a simple but effective method for searching for far away but semantically related NNs and in generalizing it to work across the board from self-supervised to semi-supervised and fully supervised settings. To summarize,

\begin{enumerate}
    \item We propose constrained mean-shift (CMSF), a generalization of MSF~\cite{koohpayegani2021mean}, to utilize additional sources of knowledge to constrain the NN search space.
    \item We develop methods to select the constraint set in self-, semi- and fully supervised settings. The retrieved samples are empirically shown to be far away in the embedding space but semantically related to the query image, providing a stronger training signal compared to MSF.
    \item CMSF achieves non-trivial gains in performance over self-supervised MSF and a direct extension of MSF to semi-supervised version. CMSF outperforms SOTA methods with comparable compute in self- and semi-supervised settings.
\end{enumerate}

\section{Method}

Similar to MSF \cite{koohpayegani2021mean}, given a query image, we are interested in pulling its embedding closer to the mean of the embeddings of its nearest neighbors (NNs). However, since top NNs are close to the target itself, they may not provide a strong supervision signal. On the other hand, far away (non-top) NNs may not be semantically similar to the target image. Hence, we constrain the NN search space to include mostly far away points with high purity. The purity is defined as the percentage of the selected NNs being from the same ground truth category as the query image. 
We use different constraint selection techniques to analyze our method in supervised, self- and semi-supervised settings.

Following MSF and BYOL, we use two embedding networks: a target encoder $f(.)$ with parameters $\theta_f$ and an online encoder $g(.)$ with parameters $\theta_g$. The online encoder is directly updated using backpropagation while the target encoder is updated as a slowly moving average of the online encoder: $\theta_f \leftarrow m\theta_f + (1-m)\theta_g$ where $m$ is close to $1$. 
We add a predictor head $h(.)$ \cite{grill2020bootstrap} to the end of the online encoder so that pulling the embeddings together encourages one embedding to be predictable by the other one and not necessarily encouraging the two embeddings to be equal. In the experiments, we use a two-layer MLP for $h(.)$.

Given a query image $x_i$, we augment it twice with transformations $T_1(.)$ and $T_2(.)$, feed them to the two encoders, and normalize them with their $\ell_2$ norm to get $u_i=\frac{f(T_1(x_i))}{||f(T_1(x_i))||_2}$ and $v_i=\frac{h(g(T_2(x_i)))}{||h(g(T_2(x_i)))||_2}$. We add $u_i$ to the memory bank $M$ and remove the oldest entries to maintain a fixed size $M$. 
We select the constraint set $C_i$ as a subset of $M$. Constraint set selection is explained in detail in Sections~\ref{sec:methods_ssl}, \ref{sec:methods_sup}, and \ref{sec:methods_semi}.
We then find the set $S_i$ of top-$k$ nearest neighbors of $u_i$ in $C_i$ including $u_i$ itself.  
Finally, we update $g(.)$ by minimizing:

\begin{equation}
    \label{eqn:eq1}
    L= \sum_{i=1}^n \frac{1}{|S_i|} \sum_{z \in S_i} v_i^Tz
    \nonumber
\end{equation}

\noindent where $n$ is the size of mini-batch and $|S_i|$ is the size of set $S_i$, \eg, $k$ in top-$k$. Finally, we update $f(.)$ with the momentum update.
In the top-$all$ variation of our method, number of neighbors $k$ is set equal to the size of $C_i$, \ie, $S_i=C_i$. Note that since $u_i$ itself is included in the nearest neighbor search, the method will be identical to BYOL \cite{grill2020bootstrap} when $k=1$ and to self-supervised mean-shift \cite{koohpayegani2021mean} when the constraint is fully relaxed ($C_i=M$). Our method covers a larger spectrum of algorithms by defining the constrained set. Below we discuss the selection of constrained set in various settings. 

\subsection{Self-Supervised Setting}
\label{sec:methods_ssl}

\begin{figure}[t]
    \centering
    \includegraphics[width=0.9\linewidth]{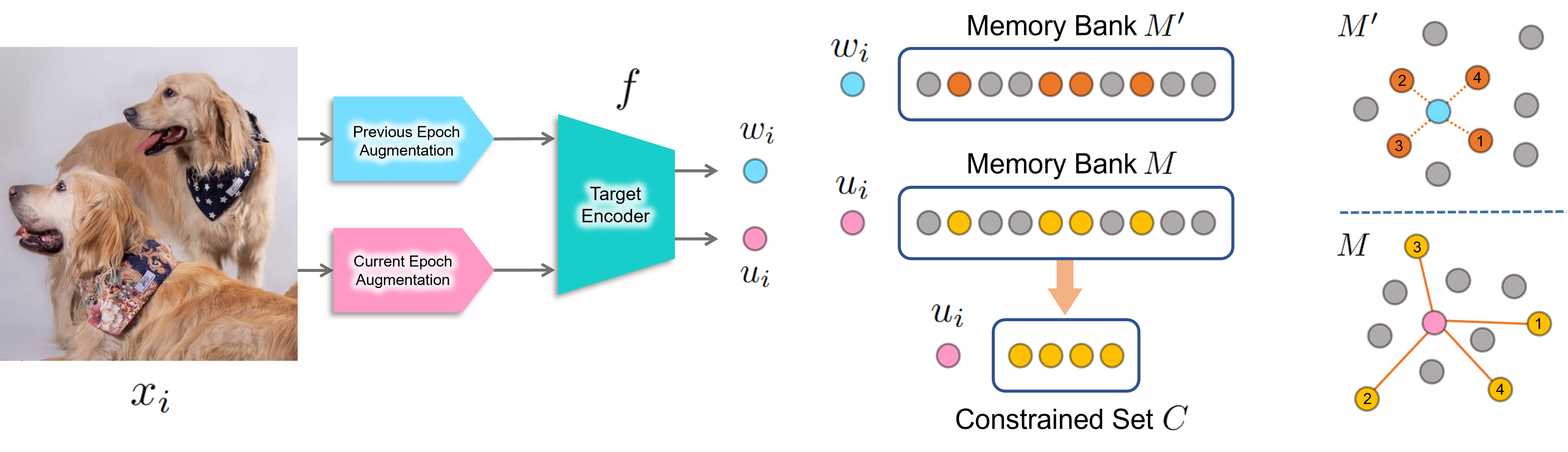}
    \caption{\textbf{$\textrm{CMSF}_{\textrm{self}}$:} The indices of the NNs of the previous epoch's memory bank $M'$ are used to construct the constrained set $C$ from the current memory bank $M$.}
    \label{fig:2q_teaser}
\end{figure}

In addition to $M$, we maintain a second memory bank $M'$ that is exactly the same as $M$ but contains features from a different ($3^{rd}$) augmentation of the image $x_i$ fed through target encoder $f(.)$. We assume $w_i\in M'$ and $u_i \in M$ are two embeddings corresponding to the same image $x_i$. Then, we find NNs of $w_i$ in $M'$ and use their indices to construct the search space $C_i$ from $M$ (See Fig.~\ref{fig:2q_teaser}). Note that although the NNs of $w_i$ in $M'$ are already close to each other, their corresponding elements in $M$ may not be close to each other since $M$ contains different augmentations $u_i$ of the same images. As a result, $C_i$ will maintain good purity while containing distant NNs (refer to Table \ref{tab:ssl_imagenet}-Right and Fig.~\ref{fig:purity}). 

Since it is expensive to embed a 3rd augmentation of each image, we embed only two augmentations as in MSF and BYOL and cache the embeddings from the previous epoch, keeping the most recent embedding for each image. The cached embedding will be still valid after one epoch since the target encoder is updated slowly using the momentum update rule (similar to MoCo). Since cache size is equal to the dataset size, we store it in the CPU memory and maintain the auxiliary memory bank $M'$ by loading the corresponding part of it to the GPU memory for each mini-batch. 
Caching of features is not essential for CMSF to work and is only used to reduce computational cost.
We performed experiments with an actual 3rd augmentation instead and found the results to be similar to our method except that it was nearly 30\% slower due to forwarding an additional augmentation. 
Table \ref{tab:ssl_imagenet}-Right shows that in the intermediate stages of learning, the top elements of $C_i$ are spread apart in $M$ with higher median ranks, and get closer to the top elements of $M$ as the learning progresses. Note that we use $w_i$ instead of $u_i$ in finding the NNs in $M'$ since both $w_i$ and $M'$ use an older target model, so are more comparable.

Since CMSF adds farther NNs only for stronger supervision, we additionally employ MSF loss calculated on the unconstrained $M$. Then, in the self-supervised setting, the total loss is an equally weighted sum of MSF and CMSF losses.

Our method can be extended to cross-modal self-supervised setting where the constraint can use NNs in a different modality rather than the 3rd augmentation of the same modality. We report the details and some preliminary experiments on this setting in the supplementary.

\subsection{Supervised Setting}
\label{sec:methods_sup}

While supervised setting is not our primary novelty or motivation, we study it to provide more insights into our constrained mean-shift framework. With access to the labels of each image, we can simply construct $C_i$ as the subset of $M$ that shares the same label as the query $x_i$. This guarantees 100\% purity for NNs. 

Note that most supervised methods, including cross-entropy loss, try to group all examples of a category together on the same side of a hyper-plane while remaining categories are on the other side. However, our method pulls the target to be close to only those examples of the same category that are already close to the target. 
This results in a supervised algorithm that may keep the latent structure of each category which can be useful for pre-training on coarse-grained labels. 
Moreover, as shown in the experiments (Fig.~\ref{fig:main_noisy}), our method is more robust to label noise since most mis-labeled images will be far from the target embedding, thus ignored in learning. This motivates applying our method to semi-supervised setting where the limited supervision provides noisy labels.

\subsection{Semi-Supervised Setting}
\label{sec:methods_semi}

In this setting, we assume access to a dataset with a small labeled and a large unlabeled subset. We train a simple classifier using the current embeddings of the labeled data and use the classifier to pseudo-label the unlabeled data. Then, similar to the supervised setting, we construct $C_i$ to be the elements of $M$ that share the pseudo-label with the target embedding. Again, this method increases the diversity of $C_i$ while maintaining high purity. To keep the purity high, we enforce the constraint only when the pseudo-label is very confident (the probability is above a threshold.) For the samples with non-confident pseudo-label, we relax the constraint resulting in regular MSF loss ({\em i.e.,} $C_i = M$.) Moreover to reduce the computational overhead of pseudo-labeling, we cache the embeddings of labeled examples throughout the epoch and train a 2-layer MLP classifier using the frozen cached features and their groundtruth labels in the middle and end of each epoch. 

\section{Experiments}

\label{sec:experiments}

\noindent {\bf Implementation details:}
We use PyTorch for all our experiments. Unless specified, we use the same hyper-parameter values in self-, semi- and fully supervised settings. All models are trained on ImageNet-1k (IN-1k) for 200 epochs with ResNet-50 \cite{he2016deep} backbone and SGD optimizer (learning rate=0.05, batch size=256, momentum=0.9, and weight decay=1e-4) with cosine scheduling for learning rate. While we focus on single crop setting in most of our experiments, we also report the results for multiple crop inputs in the SSL setting. Following SwAV~\cite{caron2020unsupervised}, we use four additional crops of $96x96$ resolution as input. These are used as inputs only to the online encoder and not the target encoder. 
The momentum value of CMSF for the moving average key encoder is $0.99$. 
The 2-layer MLP architecture for $\textrm{CMSF}_{\textrm{semi}}$ is as follows: (linear (2048x4096), batch norm, ReLU, linear (4096x512)). The default memory bank size is 128k. 
Top-$k$ is set to 10 in the semi- and fully supervised settings and 5 in the self-supervised setting. 
Additional details are provided in the supplementary. Our main CMSF experiment with 200 epochs takes nearly 6 days on four NVIDIA-2080TI GPUs. The overhead in training time due to NN search is negligible compared to the forward and backward passes through the network (that is also done in BYOL): the increase in time is 0.7\% for MSF \cite{koohpayegani2021mean} and 2.1\% for $\textrm{CMSF}_{\textrm{self}}$.

Recent SSL methods are usually computationally expensive leading to worse environmental impact and exclusion of smaller research labs. While our experiments are more efficient and accessible than most SOTA methods, \eg, PAWS, we limit our training length to 200 epochs due to resource constraints. We do not empirically verify whether the improvements observed over SOTA approaches at lower epochs (200) are persistent with longer training (\eg, 800 or 1000 epochs). 

\noindent {\bf Evaluation:}
We evaluate the pre-trained models using linear evaluation (\textit{Linear IN-1k}) in both ImageNet classification and transfer settings. The model backbone parameters are fixed and a single linear layer is trained atop them following the setting in CompRess~\cite{abbasi2020compress}. Additionally, we report $k$-nearest neighbor ($k=1, 20)$ evaluation for the SSL setting as in ~\cite{abbasi2020compress}.
The transfer performance is evaluated on the following datasets: Food101 \cite{food101}, SUN397 \cite{sun397}, CIFAR10 \cite{cifar}, CIFAR100 \cite{cifar}, Cars196 \cite{carsdataset}, Aircraft \cite{aircraft}, Flowers (Flwrs102) \cite{flowers}, Pets \cite{pets}, Caltech-101 (Calt101) \cite{caltech101}, and DTD \cite{dtd} (additional details in supplementary material.) 

\begin{table*}[!ht]
    \centering
    \caption{\textbf{Left: Evaluation on full ImageNet: } We compare our model with other SOTA methods in Linear (Top-1 Linear) and Nearest Neighbor (1-NN,20-NN) evaluation. We use a  memory bank of size 128K for CMSF and provide comparison with both 256K and 1M memory bank versions of MSF. Since $\textrm{CMSF}_{\textrm{self}}$ uses NNs from two memory banks, it is comparable to MSF (256K) in memory and computation overhead. Both single crop and multi-crop versions of our method outperform other SOTA methods, including MSF, with similar compute.
    \textbf{Right: Histogram of constrained sample ranks:} We consider the $5^{th}$ NN in the constrained set $C$ and obtain its rank in the unconstrained memory bank $M$. The histogram of these ranks are shown up to rank $100$ for different train stages of $\textrm{CMSF}_{\textrm{self}}$. Also, the median of these ranks are shown in Figure \ref{fig:purity}. A large number of distant neighbors are included in the constrained set in the early stages of training while there is a higher overlap between constrained and unconstrained NN sets towards the end of training.} 
   \begin{tabular}{cc}
   
   \scalebox{0.75}{
   
    \begin{tabular}{lcccccccc}
        \toprule
        Method & Ref. & Batch & Epochs & Sym. Loss & Multi-Crop & Top-1 & NN & 20-NN \\
        & & Size & & 2x FLOPS & Training & Linear & & \\ 
        \midrule
        Supervised & \cite{official_pytorch_models} & 256 & 100 & - & - & 76.2 & 71.4 & 74.8 \\
        \midrule
        Random-init & - & - & - & - & - & 5.1 & 1.5 & 2.0  \\
        SeLa-v2 \cite{asano2020self} & \cite{caron2020unsupervised} & 4096 & 400 & \ding{51} & \ding{55} & 67.2 & - & - \\
        SimCLR\cite{chen2020simple} & \cite{chen2020simple} & 4096 & 1000 & \ding{51} & \ding{55} & 69.3 & - & - \\
        SwAV \cite{caron2020unsupervised} & \cite{caron2020unsupervised} & 4096 & 400 & \ding{51} & \ding{55} & 70.1 & - & - \\
        DeepCluster-v2 \cite{caron2018deep} & \cite{caron2020unsupervised} & 4096 & 400 & \ding{51} & \ding{55} & 70.2 & - & - \\
        SimSiam \cite{chen2020exploring} & \cite{chen2020exploring} & 256 & 400 & \ding{51} & \ding{55} & 70.8 & - & - \\
        MoCo v2 \cite{he2020momentum} & \cite{chen2020mocov2} & 256 & 800 & \ding{55} &\ding{55} & 71.1 & 57.3 & 61.0 \\
        CompRess  \cite{abbasi2020compress} & \cite{abbasi2020compress} & 256 & 1K+130 & \ding{55} &\ding{55}& 71.9 & 63.3& 66.8\\
        InvP & \cite{wang2020invp} & 256 & 800 & \ding{55} & \ding{55} & 71.3 & - & - \\
        BYOL \cite{grill2020bootstrap} & \cite{grill2020bootstrap} & 4096 & 1000 & \ding{51} & \ding{55} & 74.3 & 62.8 & 66.9 \\
        SwAV  \cite{caron2020unsupervised} & \cite{caron2020unsupervised} & 4096 & 800 & \ding{51} & \ding{51} & 75.3 & - & - \\
        NNCLR & \cite{dwibedi2021little} & 4096 & 1000 & \ding{55} &\ding{55}& \textbf{75.4} & - & - \\
        \midrule
        SimCLR\cite{chen2020simple} & \cite{chen2020exploring} & 4096 & 200 & \ding{51} & \ding{55}& 68.3 & - & - \\
        SwAV \cite{caron2020unsupervised} & \cite{chen2020exploring} & 4096 & 200 & \ding{51} & \ding{55} & 69.1 & - & - \\
        MoCo v2 \cite{he2020momentum} & \cite{chen2020exploring} &  256 & 200  & \ding{51} & \ding{55} & 69.9 & - & - \\
        SimSiam \cite{chen2020exploring} & \cite{chen2020exploring} & 256 & 200 & \ding{51} & \ding{55} & 70.0 & - & - \\
        NNCLR\cite{dwibedi2021little} & \cite{dwibedi2021little} & 4096 & 200 & \ding{55} &\ding{55}& 70.7 & - & - \\
        BYOL \cite{grill2020bootstrap} & \cite{chen2020exploring} & 4096 & 200 & \ding{51} & \ding{55} & 70.6 & - & - \\
        SwAV \cite{caron2020unsupervised} & \cite{chen2020exploring} & 256 & 200 & \ding{51} & \ding{51} & 72.7 & - & - \\
        Truncated Triplet \cite{wang2021solving} & \cite{wang2021solving} & 832 & 200 & \ding{51} & \ding{55} & 73.8 & - & - \\
        OBoW  \cite{Gidaris_2021_CVPR} & \cite{Gidaris_2021_CVPR} & 256 & 200 & \ding{55} & \ding{51}& 73.8 & - & -\\
        \rowcolor{LightYellow}
        $\textrm{CMSF}_{\textrm{self}}$ (128K) & - & 256 & 200 & \ding{55} & \ding{51} &\bf{74.4} & \textbf{62.3} & \textbf{66.2} \\
        \rowcolor{white}
        \midrule
        MoCo v2 \cite{he2020momentum} & \cite{chen2020mocov2} & 256 & 200 & \ding{55} & \ding{55}&67.5 & 50.9 & 54.3 \\
        CO2 \cite{wei2020co2} & \cite{wei2020co2} & 256 & 200 & \ding{55} &\ding{55}& 68.0 & - & - \\
        BYOL-asym \cite{grill2020bootstrap} & \cite{koohpayegani2021mean} & 256 & 200 & \ding{55} & \ding{55} & 69.3 & 55.0 & 59.2 \\
        ISD \cite{tejankar2020isd} & \cite{tejankar2020isd} & 256 & 200 & \ding{55} & \ding{55} & 69.8 & 59.2 & 62.0 \\
        MSF (1M) \cite{koohpayegani2021mean} & \cite{koohpayegani2021mean} & 256 & 200 & \ding{55} &\ding{55}& 72.4 & 62.0 & 64.9 \\
        MSF (256K)\cite{koohpayegani2021mean} & \cite{koohpayegani2021mean} & 256 & 200 & \ding{55} &\ding{55}& 72.2 & 62.1 & 65.1 \\
        \rowcolor{LightYellow}
        $\textrm{CMSF}_{\textrm{self}}$ (128K) & - & 256 & 200 & \ding{55} & \ding{55} & \textbf{73.0} & \textbf{63.2} & \textbf{66.4} \\
        \bottomrule
        \end{tabular}}
       & 
       \includegraphics[height=0.41\textwidth,angle=0,trim=0cm 8.2cm 0 -0cm]{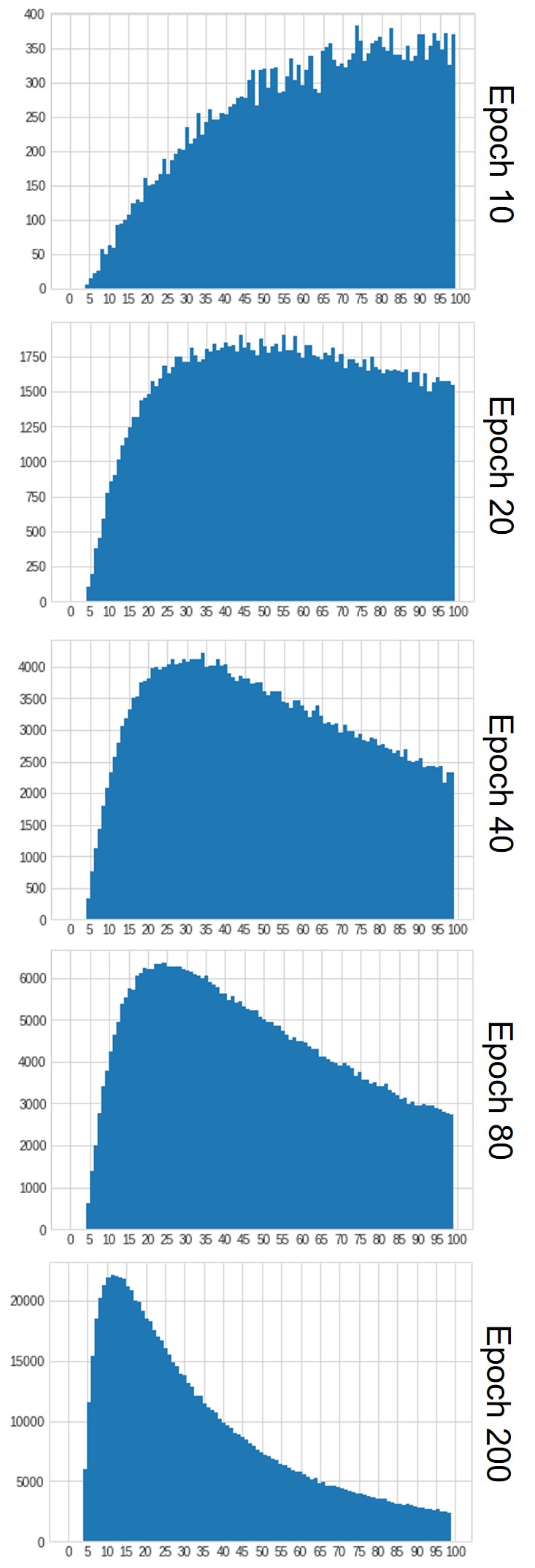} 
    \end{tabular}
\label{tab:ssl_imagenet}
\end{table*}

\begin{figure*}[t]
    \centering
    \includegraphics[width=1.0\linewidth]{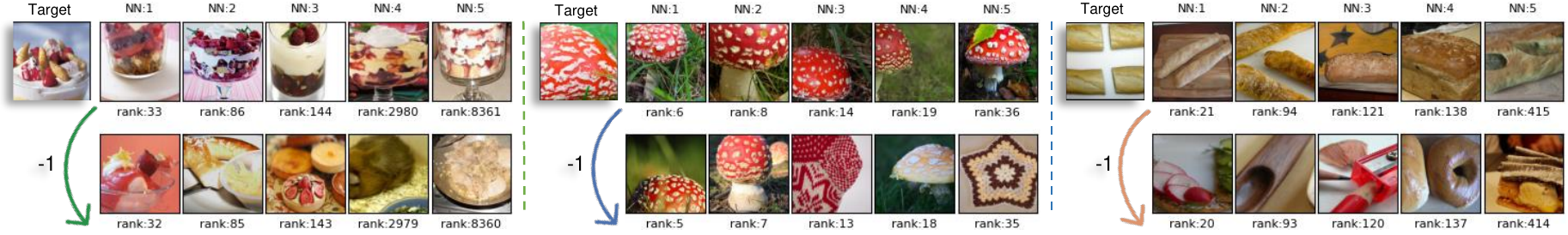}
    \caption{{\bf Nearest neighbor selection on constrained memory bank:} 
    First row shows top-5 NNs of target in constrained set $C$ and their corresponding rank in the unconstrained memory bank $M$ obtained using an intermediate checkpoint (epoch $100$). While they are not the closest samples to the target (higher rank index), they are semantically similar to the target. This shows that the constraint can capture far away samples with similar semantic as the target. The second row depicts images from memory bank with one rank lower than the corresponding image in the first row. These images contain incorrect category retrievals. Distant neighbors cannot be trivially obtained by increasing the number of NNs.
    Examples are chosen randomly.
    }
    \label{fig:visualize_nns}
\end{figure*}

\subsection{Self-Supervised Learning ($\textrm{CMSF}_{\textrm{self}}$)}

\noindent To reduce the GPU memory footprint, we cache the previous augmentation embedding of each sample in the dataset in the CPU. 
The cached features corresponding to the current mini-batch are retrieved from CPU memory to maintain memory bank $M'$ with previous augmentations. This cache is updated using the oldest features in $M$ that we remove from $M$ after each iteration.

\noindent \textbf{Results on ImageNet: } Results of $\textrm{CMSF}_{\textrm{self}}$ are shown in Table~\ref{tab:ssl_imagenet}. $\textrm{CMSF}_{\textrm{self}}$ outperforms MSF baseline with a larger memory bank, which we believe is due to pulling together far yet semantically similar samples (Fig.~\ref{fig:visualize_nns}). 
We use MSF with $2x$ larger memory bank for fair comparison. $\textrm{CMSF}_{\textrm{self}}$ also achieves state-of-the-art performance on both NN and Linear metrics when compared with approaches with similar computational budget. We compare our method to other state-of-the-art approaches with 200 epochs of training in Fig.~\ref{fig:compute_semi_sup}. We observe a good trade-off in terms of accuracy and compute for $\textrm{CMSF}_{\textrm{self}}$. Our best performance is obtained with the multi-crop version but at the cost of increased compute. \\
\noindent \textbf{Evaluation on ImageNet subsets:} Following \cite{henaff2019data,chen2020simple}, we evaluate the pre-trained models on the ImageNet classification task with limited  labels. 
We report results with 1\% and 10\% labeled subsets of ImageNet (Table~\ref{tab:lim_sup}).
$\textrm{CMSF}_{\textrm{self}}$ outperforms MSF on top-$1$ accuracy in both $1\%$ and $10\%$ settings and is comparable to existing approaches that require significantly higher training time. 

\begin{table}[t]
    \begin{center}
    \caption{\textbf{Transfer learning evaluation:} Our supervised CMSF model at just 200 epochs outperforms all supervised baselines on transfer learning evaluation. Our SSL model outperforms MSF, the comparable state-of-the-art approach, by 1.2 points on average over 10 datasets. We get the results for MoCo v2, MSF, and BYOL-asym from \cite{koohpayegani2021mean}, SimCLR and Xent (1000 epoch) from \cite{chen2020simple}, and BYOL from \cite{grill2020bootstrap}.\\}
    \label{tab:transfer_main2}
    \scalebox{0.82}{
    \begin{tabular}{|l|c|c|c|c|c|c|c|c|c|c|c||c|c|}
    \hline
    Method & \small{Epoch} & Food & \small{CIFAR} & \small{CIFAR} & SUN & Cars & Air- & DTD & Pets & Calt. & Flwr & \textbf{Mean} & \textbf{Linear} \\
    & & 101 & 10 & 100 & 397 & 196 & craft &  &  & 101 & 102 & \textbf{Trans} & \textbf{IN-1k} \\
    \hline
    \multicolumn{14}{|  c  |}{Supervised Models} \\
    \hline
    Xent & 200 & 67.7 & 89.8 & 72.5 & 57.5 & 43.7 & 39.8 & 67.9 & 91.8 & 91.1 & 88.0 & 71.0 & 77.2 \\
    Xent & 90 & 72.8 & 91.0 & 74.0 & 59.5 & 56.8 & 48.4 & 70.7 & 92.0 & 90.8 & 93.0 & 74.9 & 76.2 \\
    ProtoNW & 200 & 73.3 & 93.2 & 78.3 & 61.5 & 65.0 & 57.6 & 73.7 & 92.2 & 94.3 & 93.7 & 78.3 & 76.0 \\
    SupCon & 200 & 72.5 & 93.8 & 77.7 & 61.5 & 64.8 & 58.6 & 74.6 & \textbf{92.5} & 93.6 & 94.1 & 78.4 & \textbf{77.5} \\
    Xent & 1000 & 72.3 & 93.6 & 78.3 & 61.9 & 66.7 & 61.0 & \textbf{74.9} & 91.5 & 94.5 & 94.7 & 78.9 & 76.3 \\
    \rowcolor{LightYellow}
    $\textrm{CMSF}_{\textrm{sup}}$ top-$all$ & 200 & 73.7 & 94.2 & \textbf{78.7} & 62.1 & \textbf{71.7} & \textbf{64.1} & 73.4 & \textbf{92.5} & 94.5 & \textbf{95.8} & \textbf{80.1} & 75.7 \\
    \rowcolor{LightYellow}
    $\textrm{CMSF}_{\textrm{sup}}$ top-$10$ & 200 & \textbf{74.9} & \textbf{94.4} & \textbf{78.7} & \textbf{62.7} & 70.8 & 63.4 & 73.8 & 92.2 & \textbf{94.9} & 95.6 & \textbf{80.1} & 76.4 \\
    \hline
    \multicolumn{14}{|  c  |}{Self-Supervised Models} \\
    \hline
    
    SimCLR & 1000 & 72.8 & 90.5 & 74.4 & 60.6 & 49.3 & 49.8 & \textbf{75.7} & 84.6 & 89.3 & 92.6 & 74.0 & 69.3 \\
    MoCo v2 & 800 & 72.5 & \textbf{92.2} & 74.6 & 59.6 & 50.5 & 53.2 & 74.4 & 84.6 & 90.0 & 90.5 & 74.2 & 71.1 \\
     BYOL & 1000  & \textbf{75.3} & 91.3 & \textbf{78.4} & \textbf{62.2} & \textbf{67.8} & \textbf{60.6} & 75.5 & \textbf{90.4} & \textbf{94.2} & \textbf{96.1} & \textbf{79.2} & \textbf{74.3} \\
    
    \hline
    MoCo v2 & 200 & 70.4 & 91.0 & 73.5 & 57.5 & 47.7 & 51.2 & 73.9 & 81.3 & 88.7 & 91.1 & 72.6 & 67.5 \\
    BYOL\small{-asym} & 200 & 70.2 & 91.5 & 74.2 & 59.0 & 54.0 & 52.1 & 73.4 & 86.2 & 90.4 & 92.1 & 74.3 & 69.3 \\
    MSF & 200 & 72.3 & \textbf{92.7} & 76.3 & 60.2 & 59.4 & 56.3 & 71.7 & 89.8 & 90.9 & 93.7 & 76.3 & 72.1 \\
    \rowcolor{LightYellow}
    $\textrm{CMSF}_{\textrm{self}}$ & 200 & \textbf{73.0} & 92.2 & \textbf{77.2} & \textbf{61.0} & \textbf{60.6} & \textbf{58.4} & \textbf{74.1} & \textbf{91.1} & \textbf{92.0} & \textbf{94.5} & \textbf{77.4} & \textbf{73.0} \\
    \hline
    \end{tabular}
    }
    \end{center}
    
\end{table}

\noindent \textbf{Transfer learning:} We follow the procedure in ~\cite{grill2020bootstrap,chen2020simple} for transfer evaluation (refer to Table~\ref{tab:transfer_main2}). Hyperparameters for each dataset are tuned independently based on the validation set accuracy and final accuracy is reported on the held-out test set (more details in supplementary). $\textrm{CMSF}_{\textrm{self}}$ achieves SOTA average performance among methods trained for 200 epochs.

\begin{figure}[]
   \centering
    \includegraphics[width=0.6\linewidth]{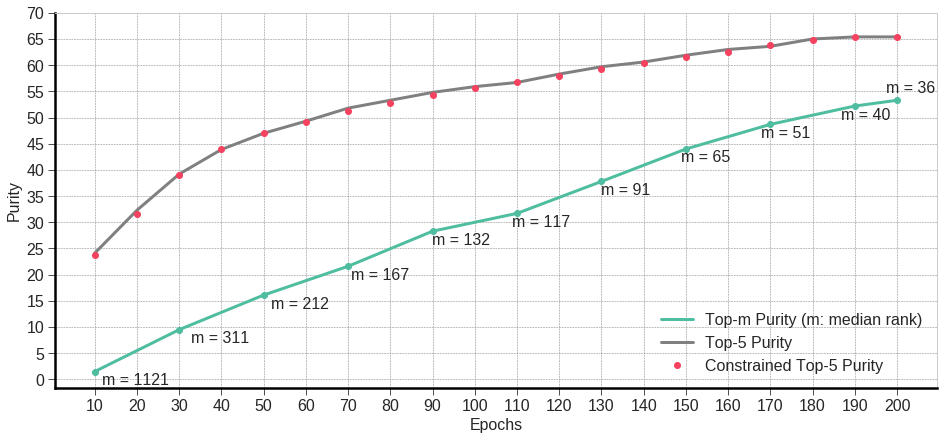}
    \caption{\textbf{Purity of constrained samples:} During training of $\textrm{CMSF}_{\textrm{self}}$ , we plot purity of the top-$5$ samples in unconstrained set $M$ (in black) and that of the top-$5$ samples in constrained set $C$ (in red). The red curve is not significantly below the black one suggesting that the purity is not dropped by increasing the distance of the NNs. To show that elements in $C$ may be far from the target $u$, we choose the $5^{th}$ element in $C$ and find its rank in the set $M$. We calculate the median of this rank as $m$.
    The purity of the top-$m$ elements of set $M$ (green curve) is consistently lower than that of top-$5$ elements of the constrained set $C$ (red curve).
    This suggests that one cannot maintain high purity by simply considering more NNs using a larger $k$. 
    }

    \label{fig:purity}

\end{figure}

\noindent \textbf{Purity of constrained samples:}
In $\textrm{CMSF}_{\textrm{self}}$, we depend on information from previous augmentations to constrain NN search in the current memory bank. Our goal is to improve learning by using distant samples with a good purity.
We observe that the top-$k$ samples from constrained memory bank $C$ have higher rank in $M$, so are far neighbors of the target (see Table \ref{tab:ssl_imagenet}-Right and Fig.~\ref{fig:purity}). Also, as shown in Fig.~\ref{fig:purity}, those samples maintain almost the same purity as the top-$k$ samples from unconstrained memory bank $M$. As a result, $C$ maintains good purity while being diverse. 

\noindent \textbf{Effect of $k'$:} In $\textrm{CMSF}_{\textrm{self}}$, we first calculate top-$k'$ samples (the first $k'$ NNs of the target) from the secondary memory bank $M'$. We then use those indices to constrain NN search space in the primary memory bank $M$ and select top-$k$ for optimization. We varied the value of $k'$ in $\textrm{CMSF}_{\textrm{self}}$ to explore its effect, keeping $k$ fixed to $5$. 
We observe that increasing $k'$ (relaxing the constraint) will decrease the accuracy of the model. As observed in Table~\ref{tab:abl_msf_topk}-right, the overlap between constrained and unconstrained NN set increases with increasing value of $k'$. Note that in a case where $k'=\infty$, $\textrm{CMSF}_{\textrm{self}}$ will be identical to the MSF baseline.

\begin{table}[t]
    \captionof{table}
    {\textbf{Effect of \textit{$k'$} in sampling NN from $M'$:}
    In $\textrm{CMSF}_{\textrm{self}}$, we constrain top-$k$ NN search space in $M$ with top-$k'$ samples from $M'$. (\textbf{Left}) Increasing $k'$ results in a drop in accuracy. The $k$ in top-$k$ is set to $5$ for all values of $k'$. (\textbf{Right}) Histogram of the constrained sample ranks at epoch $50$. The histogram shifts left, \ie, overlap between constrained and unconstrained NN sets increases with increasing value of $k'$. 
    }
    \begin{tabular}{cc}
    \scalebox{0.9}{
    \begin{tabular}{cccccc}
        \toprule
        \textit{$k'$} & 5 & 10 & 20 & 40 & 80 \\
        \midrule
        NN & 63.2 & 62.9 & 62.7 & 62.3 & 61.7 \\
        20-NN & 66.4 & 66.1 & 65.9 & 65.6 & 65.0 \\
        \bottomrule
    \end{tabular}
    }
    &
    \includegraphics[width=0.66\linewidth,angle=0,trim=0cm 1.5cm 0 -0cm]{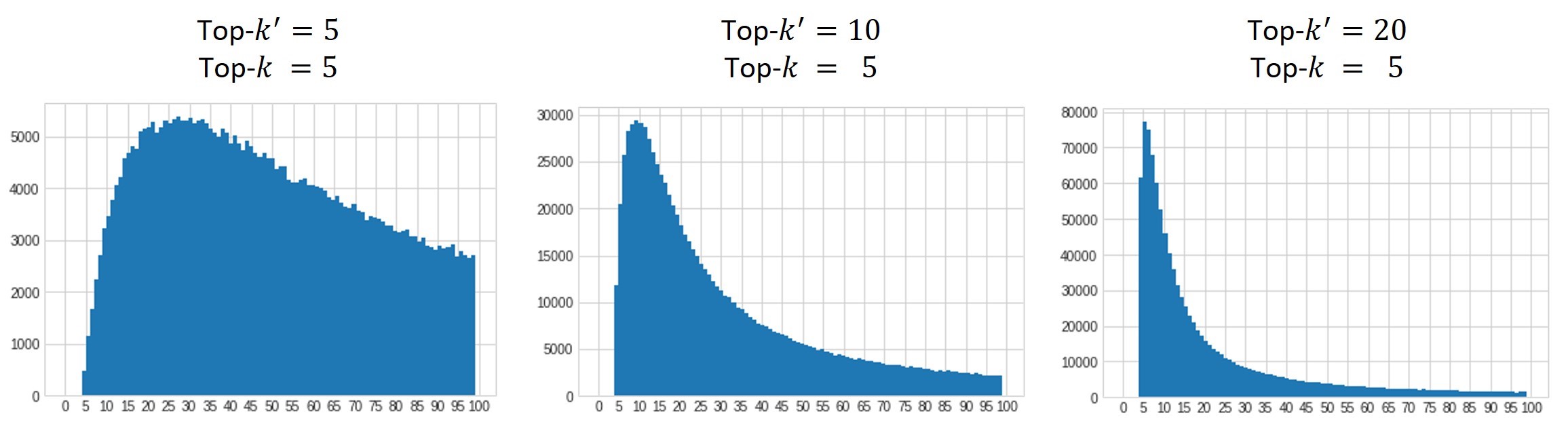}
    \end{tabular}
    \label{tab:abl_msf_topk}
\end{table}

\subsection{Supervised Learning}
\label{sec:supervised}

\noindent \textbf{Evaluation:} Unlike cross-entropy (Xent~\cite{NIPS1987_eccbc87e,levin1988accelerated,rumelhart1986learning}) baseline, SupCon~\cite{khosla2020supervised}, ProtoNW~\cite{snell2017prototypical} and CMSF do not train a linear classifier during the pre-training stage. Thus, we use the pre-training dataset ImageNet-1k (IN-1k) for linear evaluation of the frozen features as done in SSL. 
For Xent, we use the linear classifier trained during pre-training. We use the same settings and datasets as self-supervised for transfer learning evaluation.

\begin{figure}[t]
\centering
\begin{minipage}{0.50\linewidth}
    \captionof{table}{\textbf{Evaluation on small labeled ImageNet} : We compare our model to MSF and other baselines on ImageNet 1$\%$ and 10$\%$ linear evaluation benchmarks. ``Fine-tuned'' refers to fine-tuning the entire backbone network instead of a single linear layer. $\textrm{CMSF}_{\textrm{self}}$ outperforms MSF on top-1 metric in both $1\%$ and $10\%$ settings.}
    \scalebox{0.8}{
    \begin{tabular}{lcccccc}
        \toprule
        \multirow{2}{*}{Method} & Fine- & \multirow{2}{*}{Epochs} & \multicolumn{2}{c}{Top-1} & \multicolumn{2}{c}{Top-5} \\
        & tuned & & 1\% & 10\% & 1\% & 10\% \\
        \midrule
        Supervised & \ding{51} & & 25.4 & 56.4 & 48.4 & 80.4 \\
        PIRL \cite{misra2019self} & \ding{51} & 800 & - & - & 57.2 & 83.8  \\
        CO2 \cite{wei2020co2} & \ding{51} & 200 & - & - & 71.0 & 85.7 \\
        SimCLR \cite{chen2020simple} & \ding{51} & 1000 & 48.3 & 65.6 & 75.5 & 87.8 \\
        InvP \cite{wang2020invp} & \ding{51} & 800 & - & - & 78.2 & 88.7 \\
        BYOL \cite{grill2020bootstrap} & \ding{51} & 1000 & 53.2 & 68.8 & 78.4 & 89.0 \\
        $\text{SwAV}$ \cite{caron2020unsupervised} & \ding{51} & 800 & \textbf{53.9} & \textbf{70.2} & \textbf{78.5} & \textbf{89.9} \\
        \midrule
        MoCo v2 \cite{chen2020mocov2} & \ding{55} & 800 & 51.5 & 63.6 & 77.6 & 86.1 \\
        $\text{BYOL}$ \cite{grill2020bootstrap} & \ding{55} & 1000 & 55.7 & \textbf{68.6} & 80.0 & \textbf{88.6} \\
        CompRess \cite{abbasi2020compress} & \ding{55} & 1K+130 & \textbf{59.7} & 67.0 & \textbf{82.3} & 87.5 \\%[0.1cm]
        \midrule
        MoCo v2 \cite{chen2020mocov2} & \ding{55} & 200 & 43.6 & 58.4 & 71.2 & 82.9 \\
        BYOL-asym  & \ding{55} & 200 & 47.9 & 61.3 & 74.6 & 84.7 \\
        ISD \cite{tejankar2020isd} & \ding{55} & 200 & 53.4 & 63.0 & 78.8 & 85.9 \\
        MSF \cite{koohpayegani2021mean} & \ding{55} & 200 & 55.5 & 66.5 & \textbf{79.9} & 87.6 \\
        \rowcolor{LightCyan}
        \rowcolor{LightCyan}
        $\textrm{CMSF}_{\textrm{self}}$  & \ding{55} & 200 & \textbf{56.4} & \textbf{67.5} & 79.8 & \textbf{87.7} \\
        \bottomrule
    \end{tabular}
    }
    \label{tab:lim_sup}
\end{minipage}
\hfill
\begin{minipage}{0.47\linewidth}
    \centering
    \includegraphics[width=.8\linewidth]{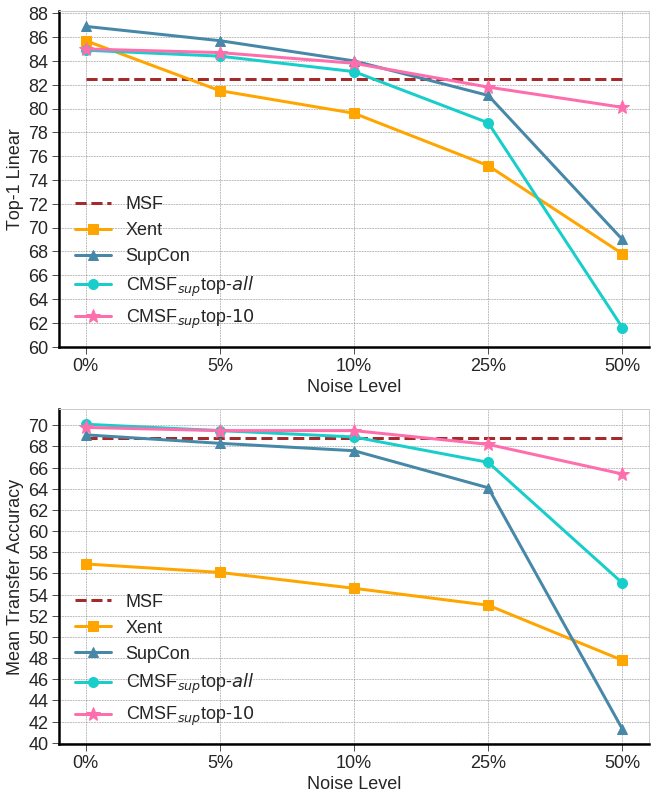}
    \caption{\textbf{Noisy supervised setting on ImageNet-100:} Our method is more robust to noisy annotation compared to Xent and SupCon. Also, using top-$all$ degrades the results since all images from a single category are not guaranteed to be semantically related due to noisy labels. Mean Transfer Accuracy is the average over 10 transfer datasets.}
   \label{fig:main_noisy}
\end{minipage}
\end{figure}

\noindent \textbf{Results:}
Results on IN-1k dataset are shown in Table \ref{tab:transfer_main2}.  In top-$all$ variation of our method, $k$ is equal to the total size of $C$. SSL inspired methods like CMSF and SupCon significantly outperform Xent when trained for similar number of epochs. We observe that improvements in ImageNet performance do not always translate to transfer performance. Interestingly, CMSF performs the best on transfer evaluation, particularly on fine-grained datasets like Cars196 and Aircraft. We believe that the absence of explicit cross-entropy based optimization using the supervised labels preserves the multi-modal distribution of categories improving fine-grained performance. Supervised CMSF uses class labels only as a constraint for MSF during pre-training and does not explicitly optimize on the classification task. Superior performance of $\textrm{CMSF}_{\textrm{sup}}$ top-$10$ demonstrates the importance of using distant yet semantically related neighbors as positives. 

\noindent \textbf{Noisy Labels: } In the noisy setting, we use random i.i.d. noise to corrupt the labels (change the label randomly) of a percentage of images. We consider, 5\%, 10\%, 25\%, and 50\% label corruption (noise) rates.  For faster experiments, we report results on the ImageNet-100 dataset~\cite{tian2019cmc} (Fig.~\ref{fig:main_noisy}). We observe a significantly higher degradation in performance of Xent baseline and $\textrm{CMSF}_{\textrm{sup}}$ top-$all$ compared to $\textrm{CMSF}_{\textrm{sup}}$ top-$10$ at high noise levels. The gap between the approaches is larger on transfer learning. 
These observations indicate that NN based methods like CMSF are better suited for noisy constraint settings compared to approaches utilizing all samples of a class as positives. This robustness to label noise motivates our application of CMSF to self- and semi-supervised settings where pseudo-labels or the NNs of previous augmentations may be noisy.

\noindent \textbf{Coarse-grained ImageNet:} CMSF groups together only top-$k$ neighbors and thus can help in preserving the latent structure of the data compared to top-$all$. To verify this, we consider a dataset with coarse-grained labels where this difference is pronounced. Based on the WordNet hierarchy, we merge each category in the ImageNet dataset to its parent class. We further ensure that no two classes are in the same path in the graph by merging the descendant into the ancestor class. The total number of classes is thus reduced from $1000$ in ImageNet-1k to $93$ in our ImageNet-coarse. We train CMSF and the baseline approaches in a supervised manner using the coarse labels and then evaluate on the fine-grained / original labels on ImageNet-1k validation set. In Table~\ref{tab:coarse_imagenet} we compare the top-$all$ and top-$k$ variants on the coarse grained version of ImageNet. $\textrm{CMSF}_{\textrm{sup}}$ top-$k$ sees a minor drop in performance compared to training on ImageNet-1k. However, methods in which all samples in a class are explicitly brought closer - $\textrm{CMSF}_{\textrm{sup}}$ top-$all$, cross-entropy and supervised contrastive - see a huge drop in accuracy. More details on coarse-grained ImageNet are in the supplementary.

\setlength{\tabcolsep}{4pt}
\begin{table}[t]
    \centering
    \caption{\textbf{Supervised learning on coarse grained ImageNet:} We train on the coarse grained version of ImageNet (93 super categories) and perform linear evaluation on the original ImageNet-1k validation set with fine-grained labels (1000 categories).}
    \scalebox{0.85}{
    \begin{tabular}{ccccc}
         \toprule
         \multirow{2}{*}{Train Dataset}                    & \multicolumn{4}{c}{ImageNet-1k Validation Set} \\
                & Xent & SupCon & $\textrm{CMSF}_{\textrm{sup}}$ top-$all$ & $\textrm{CMSF}_{\textrm{sup}}$ top-$10$ \\
         \midrule
         ImageNet-1k         & 77.2 & \textbf{77.5}   & 75.7         & 76.4 \\
         ImageNet-coarse     & 61.4 & 58.7   & 67.0         & \textbf{74.2} \\
    \bottomrule
    \end{tabular}}
    \label{tab:coarse_imagenet}
\end{table}

\subsection{Semi-Supervised Learning}

\noindent \textbf{Implementation Details:} We train a 2-layer MLP atop the cached target features of supervised set for pseudo-labeling. The pseudo-label training is performed twice per epoch (takes 40 seconds per training) and the label assignment is done in an online fashion for each mini-batch. The confidence threshold for pseudo-labeling is set to 0.85. We use the same optimizer settings as in self-supervised CMSF for the pre-training stage. Similar to S4L~\cite{Zhai_2019_ICCV}, we perform two stages of fine-tuning with supervised and pseudo-labels. We fine-tune the backbone network with two MLPs (as in PAWS~\cite{assran2021semi}) on the 10\% labeled set for 20 epochs 
and pseudo-label the train set. Samples above confidence threshold (nearly 30\% of dataset) are combined with supervised set to fine-tune again for 20 epochs  (more details in suppl.). 
The second fine-tuning is equivalent to 5 epochs with full data and is a small increase in our total compute. This is needed since we do not directly optimize cross-entropy loss in pre-training as in ~\cite{sohn2020fixmatch,xie2019unsupervised,pham2021meta}.

\noindent \textbf{Evaluation:} The final epoch parameters are used to perform evaluation. We report top-1 accuracy on the ImageNet validation set. We additionally report the total number of FLOPs for forward and backward passes (backward is $2\times$ forward) through ResNet-50 backbone and the number of GPUs/TPUs used by each method in the pre-training stage (more details in suppl.). 

\setlength{\tabcolsep}{3pt}
\begin{table}[t]
    \centering
    \caption{\textbf{Semi-supervised learning on ImageNet dataset with 10\% labels:} FLOPs denotes the total number of FLOPS for forward and backward passes through ResNet-50 backbone while batch size denotes the sum of labeled and unlabeled samples in a batch. $\textrm{CMSF}_{\textrm{semi}}$-mix precision is compute and resource efficient, achieving SOTA performance at comparable compute. PAWS requires large number of GPUs to be compute efficient and its performance drastically drops with 4/8 GPUs. $^{\dagger}$ Trained with stronger augmentations like RandAugment~\cite{NEURIPS2020_d85b63ef}. * TPUs are used.}
    \scalebox{0.8}{
    \begin{tabular}{lccccc}
        \toprule
        Method & Epochs  &  Batch & GPUs & FLOPs    & Top-1 \\
               &         &  Size  &      & (x$10^{18}$) &       \\
        \midrule
        \multicolumn{4}{l}{\textit{Self-supervised Pre-training}} \\
        Mean Shift~\cite{koohpayegani2021mean}          & 200  & 256   & 4   & 4   & 67.4  \\        
        BYOL~\cite{grill2020bootstrap}                  & 1000 & 4096  & 512*& 40  & 68.8  \\        
        SwAV~\cite{caron2020unsupervised}               & 800  & 4096  & 64  & 37  & 70.2  \\        
        SimCLRv2~\cite{chen2020big}                     & 800  & 4096  & 128*& 16  & 68.4  \\         
        \midrule                                                                                
        \multicolumn{4}{l}{\textit{Semi-supervised Pre-training}} \\                            
        SimCLRv2 (+Self Dist)~\cite{chen2020big}        & 1200 & 4096  & 128*& 20  & 70.5  \\        
        UDA$^{\dagger}$~\cite{xie2019unsupervised}      & 800  & 15872 & 64* & 10  & 68.1  \\  
        FixMatch$^{\dagger}$~\cite{sohn2020fixmatch}    & 300  & 6144  & 32* & 7   & 71.5  \\  
        MPL$^{\dagger}$~\cite{pham2021meta}             & 800  & 2048  & -   & 30  & 73.9  \\  
        PAWS (support=6720)~\cite{assran2021semi}       & 300  & 4096  & 64  & 21  & 75.5 \\    
        PAWS (support=1680)~\cite{assran2021semi}       & 100  & 256   & 8   & 15  & 70.2 \\      
        PAWS (support=400)~\cite{assran2021semi}        & 100  & 256   & 4   & 7   & 62.9 \\      
        \rowcolor{LightYellow}                                                                 
        $\textrm{CMSF}_{\textrm{semi}}$-basic        & 200  & 256   & 4   & 4   & 68.6  \\        
        \rowcolor{LightYellow}                                                                 
        $\textrm{CMSF}_{\textrm{semi}}$                 & 200  & 256   & 4   & 4   & 69.9  \\  
        \rowcolor{LightYellow}                                                                 
        $\textrm{CMSF}_{\textrm{semi}}$-mix precision   & 200  & 768   & 4   & 4   & 70.5  \\  
        \bottomrule
    \end{tabular}
    }
    \label{tab:semi_sup_1p10p}
    \end{table}
\noindent \textbf{Baselines:} We compare the proposed approach (\textit{$\textrm{CMSF}_{\textrm{semi}}$}) with self- and semi-supervised approaches. \textit{$\textrm{CMSF}_{\textrm{semi}}$-basic} minimizes unconstrained MSF loss on the unlabeled examples (no pseudo-labeling) and CMSF loss on the labeled examples only. We provide comparison of PAWS method with different support set sizes. We train PAWS on 4x 16GB GPUs with maximum possible support set size (200 classes, 2 images/class) using code provided by the authors. We also report results using mixed precision training (\textit{$\textrm{CMSF}_{\textrm{semi}}$-mix precision}) as in PAWS~\cite{assran2021semi} with a higher batch size of 768 since it has lower memory requirement. 

\noindent \textbf{Results:} $\textrm{CMSF}_{\textrm{semi}}$-mix precision achieves comparable performance to most methods with significantly less training and without the use of stronger augmentation schemes like RandAugment~\cite{NEURIPS2020_d85b63ef} (Table~\ref{tab:semi_sup_1p10p}, Fig.~\ref{fig:compute_semi_sup}). PAWS with a support set size of 6720 outperforms other approaches. However, this requires significantly higher compute (4.8$\times$ FLOPs) and resources (64 GPUs) compared to $\textrm{CMSF}_{\textrm{semi}}$-mix precision (4 GPUs). Since PAWS requires a large support set, it does not scale well to lower resource (4/8 GPUs) settings even if the total compute remains the same. When trained on only 4 GPUs, CMSF outperforms PAWS by \textbf{7.6\%} points. Additional ablations and results on ImageNet-100 dataset are in supplementary.

\section{Related Work}

\noindent {\bf Self-supervised learning (SSL): } 
Earlier works on SSL focused on solving a pretext task  that does not require additional labeling.
Examples of pretext tasks include colorization \cite{zhang2016colorful}, jigsaw puzzle~\cite{noroozi2016unsupervised}, counting~\cite{noroozi2017representation}, and rotation prediction~\cite{gidaris2018unsupervised}.  Another class of SSL methods is based on instance discrimination \cite{dosovitskiy2014discriminative}. The idea is to classify each image as its own class. Some methods adopt the idea of contrastive learning for instance discrimination \cite{he2020momentum,chen2020simple,caron2018deep,caron2020unsupervised,caron2021emerging}. BYOL \cite{grill2020bootstrap} proposes a non-contrastive approach by removing the negative set and simply regressing one view of an image from another. 

Several recent works aim to find a larger positive sample set to improve learning. In LA~\cite{zhuang2019local}, samples are clustered using $k$-means and samples within a cluster are brought closer together compared to cross-cluster samples. 
MSF \cite{koohpayegani2021mean} and MYOW~\cite{azabou2021mine} generalize BYOL by regressing target view and its NNs. NNCLR \cite{dwibedi2021little} extends SimCLR to use NNs as positives. CLD \cite{Wang_2021_CVPR} integrates grouping using instance-group discrimination.  
Affinity diffusion~\cite{huang2020unsupervised} uses strongly connected nodes in a graph constructed using embeddings to find positive samples. 
Unlike these methods, we focus on grouping together far away neighbors that are semantically similar. We show quantitatively and qualitatively the diversity and purity of retrieved neighbors and improved performance over MSF. We generalize the idea in MSF \cite{koohpayegani2021mean} to use an additional source of knowledge to constrain the NN search space for the target view.
CoCLR~\cite{han2021selfsupervised} and Cl-InfoNCE~\cite{tsai2021integrating} also use additional information sources in the form of additional modality and auxiliary labels respectively to improve performance.  
However, we focus on self- and semi-supervised classification settings and design methods to obtain and use the additional information as a constraint in NN search space. 

\noindent {\bf Supervised learning:} 
A drawback of Cross-entropy is its lack of robustness to noisy labels \cite{zhang2018generalized,sukhbaatar2015training}. 
\cite{szegedy2015rethinking,muller2020does,touvron2020grafit,xu2021weakly} address the issue of hard labeling, \eg, (one-hot labels) with label smoothing, \cite{hinton2015distilling,bagherinezhad2018label,furlanello2018born} replace hard labels with prediction of pre-trained teacher, and \cite{zhang2018mixup,yun2019cutmix} propose an augmentation strategy to train on combination of instances and their labels. Another line of work \cite{goldberger2004neighbourhood,salakhutdinov2007learning} is to learn representations with good kNN performance. SupCon~\cite{khosla2020supervised} and \cite{wu2018improving} improve upon \cite{goldberger2004neighbourhood} by changing the distance to inner product on $\ell_2$ normalized embeddings. 
We include the supervised setting to better understand the effect of using constrained NNs, particularly in the noisy label setting. 

\noindent {\bf Semi-supervised learning:} Several methods combine self-supervised and supervised learning to form semi-supervised methods. S4L~\cite{Zhai_2019_ICCV} uses rotation prediction based loss on the unlabeled set along with cross-entropy loss on the labeled set. Similarly, SuNCEt~\cite{assran2020supervision} combines SimCLR~\cite{chen2020simple} and SwAV~\cite{caron2020unsupervised} methods with supervised contrastive loss. Pseudo-labeling is frequently used in semi-supervised learning. In Pseudo-Label~\cite{lee2013pseudo}, the network is trained with cross-entropy loss using supervised data on the labeled examples and pseudo-labels on the unlabeled ones. In SimCLR-v2~\cite{chen2020big}, a teacher network is pre-trained using SimCLR~\cite{chen2020simple} and fine-tuned with supervised labels. The teacher is then distilled to a student network using pseudo-labels on the unlabeled set. FixMatch~\cite{sohn2020fixmatch} uses pseudo-labels obtained using a weakly augmented image to train a strongly augmented version of the same image. UDA~\cite{xie2019unsupervised} leverages strong data augmentation techniques in enforcing this consistency in pseudo-labels across augmentations. MPL~\cite{pham2021meta} optimizes a student network using pseudo-labels from a teacher network, while the teacher is optimized to maximize the student's performance on the labeled set. PAWS~\cite{assran2021semi} uses consistency based loss on soft pseudo-labels obtained in a non-parametric manner. Our method too uses pseudo-labels to train the unlabeled samples. 
However, we use the labels as a constraint in MSF~\cite{koohpayegani2021mean} and do not directly optimize samples using cross-entropy loss.

\noindent {\bf Metric learning:} 
The goal of metric learning is to train a representation that puts two instances close in the embedding space if they are semantically close. 
Two important methods in metric learning are: triplet loss \cite{chopra2005learning,weinberger2006distance,schroff2015facenet} and contrastive loss \cite{sohn2016improved,bromley1993signature}. 
Metric learning methods perform well on tasks like image retrieval \cite{Wu_2017_ICCV} and few-shot learning \cite{vinyals2017matching,snell2017prototypical}. Prototypical networks \cite{snell2017prototypical} is similar to a contrastive version of our method with top-$all$.

\section{Conclusion}
MSF is a recent SSL method that pulls an image towards its nearest neighbors. We argue that the model can benefit from more diverse yet pure neighbors. Hence, we generalize MSF method by constraining the NN search. This opens the door to using the mean-shift idea to various settings of self-supervised, supervised, and semi-supervised. To construct the constraint, our SSL method uses cached  augmentations from the previous epoch while the supervised and semi-supervised settings use labels or pseudo-labels. We show that our method outperforms SOTA approaches like MSF in SSL, PAWS in semi-supervised, and supervised contrastive in transfer-learning evaluation of supervised settings.

\textbf{Acknowledgments:}
This material is based upon work partially supported by DARPA under Contract No. HR00112190135, the United States Air Force under Contract No. FA8750‐19‐C‐0098, funding from SAP SE, and NSF grants 1845216 and 1920079. Any opinions, findings, and conclusions or recommendations expressed in this material are those of the authors and do not necessarily reflect the views of the United States Air Force, DARPA, or other funding agencies. 

% %%%%%%%%% REFERENCES
\newpage
\bibliographystyle{splncs04}
\bibliography{egbib}
\end{document}

\pagestyle{headings}
\mainmatter
\def\ECCVSubNumber{4300}  % Insert your submission number here

\title{Constrained Mean Shift Using Distant Yet \\ Related Neighbors for Representation Learning: Supplementary Material}

%******************

\titlerunning{Constrained Mean Shift Using Distant Yet Related Neighbors}
\author{K L Navaneet\inst{1} $^{*}$ \index{Navaneet, K L} \and Soroush Abbasi Koohpayegani\inst{1} $^{*}$ \index{Abbasi Koohpayegani, Soroush} \and Ajinkya Tejankar\inst{1} $^{*}$\and Kossar Pourahmadi\inst{1}\and Akshayvarun Subramanya\inst{2} \and Hamed Pirsiavash\inst{1}
 }
\authorrunning{Navaneet, Abbasi Koohpayegani, Tejankar et al.}

 \institute{
University of California, Davis \and University of Maryland, Baltimore County 
}
%******************
\maketitle

Here, we provide additional results and analysis on self-supervised (section~\ref{results_ssl}), cross modal constraint (section~\ref{results_cross}), semi-supervised (section~\ref{results_semisup}) and supervised (section~\ref{results_sup}) settings. Additional results include analysis of retrieved far neighbors (Figs.~\ref{fig:semi_sup_purity} and ~\ref{fig:vis_nn_2q}) and ablations to justify various design choices (tables~\ref{tab:conf_th}, ~\ref{tab:pseudo_cache_schemes}, ~\ref{tab:mlp_arch}, ~\ref{tab:fine_tune}, ~\ref{tab:all_ablations}, ~\ref{tab:main_noisy}, ~\ref{tab:appendix_transfer_dset_details}).  More details on implementation (section~\ref{impl_details}) and compute calculation (Eq.~\ref{eqn:comp_calc}, table~\ref{tab:flops_calc_semi_sup}) are provided. We also provide the code as part of the supplementary materials.

\section{Results on Self-supervised Setting}
\label{results_ssl}

\subsection{Using far neighbors in unconstrained MSF: } In CMSF$_{\textrm{self}}$, we use augmented images from previous epoch to obtain distant neighbors. A trivial way to sample farther neighbors is to just increase the number of neighbors $k$ in the original (unconstrained) MSF~\cite{koohpayegani2021mean} method. Here we train MSF with $k=500$ to compare with our $\textrm{CMSF}_{\textrm{self}}$. We train all models for 80 epochs. Settings are similar to our self-supervised settings in section 3.1. For fair comparison, we use memory-bank of size $256k$ for MSF while we use $128K$ for $\textrm{CMSF}_{\textrm{self}}$. Results are in Table \ref{tab:abl_msf_hightopk}. While increasing $k$ in MSF helps to sample far NNs, it degrades the accuracy. We hypothesize that this happens due to reducing purity of top-$k$ in unconstrained MSF with increasing $k$ (also shown in Fig. 5 of the main submission). This experiment shows that it is not trivial to to sample far NNs with good purity.  

\begin{table}[h!]
    \begin{center}
    \caption{\textbf{Using far neighbors in unconstrained MSF~\cite{koohpayegani2021mean}:} Using far NNs by trivially increasing $k$ in MSF baseline degrades the accuracy. This is due to the low purity of far NNs in MSF when no constraint is utilized. However, CMSF$_{\textrm{self}}$ achieves high accuracy while using distant NNs.}
    \scalebox{1.0}{
    \begin{tabular}{cC{3cm}C{3cm}C{3cm}}
        \toprule
        & MSF~\cite{koohpayegani2021mean} & MSF~\cite{koohpayegani2021mean} & $\textrm{CMSF}_{\textrm{self}}$  \\
        & Top-$k=500$ & Top-$k=5$ & Top-$k=k'=5$  \\
        \midrule
        NN & 35.8 & 49.7 & 51.4\\
        20-NN & 40.2 & 54.0 & 55.5 \\
        \bottomrule
    \end{tabular}
    }
    \end{center}
    \label{tab:abl_msf_hightopk}
\end{table}

\subsection{Effect of number of neighbors $k$} 

$\textrm{CMSF}_{\textrm{self}}$ uses top-$k$ NNs as part of loss calculation. Here we study the effect of $k$ in $\textrm{CMSF}_{\textrm{self}}$ performance. We set $k'=k$ in all models. Note that as described in Line 350 of the main submission, $k'$ is the number of NNs retrieved from the second memory bank $M'$. We train all models for 80 epochs. All settings are similar to that of $\textrm{CMSF}_{\textrm{self}}$ in Section 3.1. Results are shown in Table \ref{tab:abl_cmsf_topk}. Higher values of $k$ and $k'$ degrade model performance. We thus use $k'=k=5$ in all our main experiments.  

\begin{table}[h!]
    \begin{center}
    \caption{\textbf{Effect of \textit{$k$} in top-\textit{$k$} NNs sampling within $M$:}  We set $k=k'$ in all models and varied $k$. We use memory bank of size $256k$. Increasing $k$ degrades the accuracy of the model. }
    \scalebox{1.0}{
    \begin{tabular}{ccccc}
        \toprule
        & \textit{$k'$}=\textit{$k$}=5 \quad & \textit{$k'$}=\textit{$k$}=10 \quad& \textit{$k'$}=\textit{$k$}=20 \quad& \textit{$k'$}=\textit{$k$}=50 \\
        \midrule
        NN & 51.4 & 51.3 & 51.1 & 49.5  \\
        20-NN & 55.5 & 55.3 & 55.3 & 53.8  \\
        \bottomrule
    \end{tabular}
    }
    \end{center}
    \label{tab:abl_cmsf_topk}
\end{table}

we experimented with ResNet-18 and ResNet-101 networks. ResNet-101 is trained for 100 epochs. CMSF outperforms MSF by 1.9 points (51.7\% vs 49.8\%) with ResNet-18 and 0.8\% points (71.9\% vs 71.1\%) with ResNet-101.

\subsection{Results with Different Architectures}
In addition to the ResNet-50 architecture, we experiment with a smaller and a larger backbone architecture. We consider ResNet-18 and ResNet-101 networks. The results are shown in table~\ref{tab:arch}. The proposed CMSF$_{\textrm{self}}$ improves over MSF across different architectures. Note that the networks are trained only for $100$ epochs on ResNet-101.

\begin{table}[]
    \centering
    \caption{\textbf{Results with different backbone architectures:} We compare performance of our method with that of CMSF with ResNet architectures of different sizes. We observe that CMSF consistently outperforms MSF. * models were trained for 100 epochs instead of 200.}
    \begin{tabular}{cccc}
        \toprule
        Method & ResNet-18 & ResNet-50 & ResNet-101* \\
        \midrule
        MSF~\cite{koohpayegani2021mean} & 49.8 & 72.2 & 71.1 \\
        CMSF$_{\textrm{self}}$ & 51.7 & 73.0 & 71.9 \\
        \bottomrule 
    \end{tabular}
    \label{tab:arch}
\end{table}

\section{Cross-modal Constraint }
\label{results_cross}
Fig. 6 of main submission showed that proposed CMSF$_{\textrm{sup}}$ is more robust to noisy labels. Here, we explore another such noisy constraint: a pre-trained SSL model from another modality. 
We consider an unlabeled video dataset and use the RGB and optical flow inputs as the two different modalities. We first train two SSL models on the RGB and Flow modalities separately using InfoNCE method \cite{oord2018representation,han2021selfsupervised}. Then we continue the training on one modality while freezing the other modality and using it as a constraint. In training the flow network using RGB network as constraint, we sample $k'$ nearest neighbors in RGB's memory bank and then search for top-$k$ nearest neighbors among those samples in the memory bank corresponding to Flow. \\

\noindent \textbf{Implementation Details.}
Following \cite{han2021selfsupervised}, we use split-1 of UCF-101 \cite{soomro2012ucf101} (13k videos) as the unlabeled dataset.
We use similar augmentation and pre-processing as \cite{han2021selfsupervised} and calculate optical-flow using unsupervised TV-L1 \cite{Zach2007ADB} algorithm.
For cross-modal experiments, we use S3D \cite{xie2018rethinking} architecture with the input size of $128 \times 128$ pixels. We initialize from the pretrained weights of InfoNCE (400-epoch) released by \cite{han2021selfsupervised}. We use following settings for our method: memory bank of size $8192$, $n=10$, $k=5$, batch size $128$, weight decay $1e-5$, initial lr of $0.001$, and learning rate decay by factor of $10$ at epoch 80. We train each modality for additional 100 epochs using PyTorch Adam optimizer. For a fair comparison, we run CoCLR using their official code by initializing it from the same model as ours.
We use the code from \cite{han2021selfsupervised} for linear evaluation.  \\

\noindent \textbf{Results:}
The results are shown in Table \ref{tab:co_training}. We report top-1 accuracy for linear classification and recall@1 for retrieval on the extracted features of frozen networks. All experiments use spatio-temporal 3D data either in RGB or flow format. At the end of 3 stages of training on Flow modality, our method outperforms CoCLR~\cite{han2021selfsupervised} baseline and MSF with 2 stages. 

\begin{table}[h!]
    \centering
    \caption{{\bf Cross-modal constraint: } 
    We initialize all models using an InfoNCE pre-trained model. In CMSF-cross modal, one of the modalities is used to constrain and train the other.
    Superscript indicate the constraint modality, subscript indicate the training modality. For example, in $\textrm{CMSF}_{\textrm{RGB}}^{\textrm{Flow}}$, we continue training CMSF on RGB modality while using frozen pretrained Flow model as the constraint. Note that CoCLR \cite{han2021selfsupervised} also uses another modality as a constraint in the form of contrastive learning. 
    We continue training InfoNCE SSL model for 200 epochs using MSF \cite{koohpayegani2021mean} for a fair comparison. 
    We use S3D \cite{xie2018rethinking} architecture for all models. Models with the final round of training on Flow modality are highlighted with yellow and those on RGB are highlighted with blue. All rows with * contain results for the same model. Results are repeated for easier understanding of the table.}
    \label{tab:co_training}
    \scalebox{0.9}{
    \begin{tabular}{lC{2cm}C{3cm}cc}
        \toprule
        Model  &  Final & Epochs & R@1 & Linear \\
               & modality & & & \\ 
        \midrule
        \rowcolor{LightCyan}
        $\textrm{InforNCE}_{RGB}$  & RGB & 400 & 35.5 & 47.9 \\
        \rowcolor{LightCyan}
        $\textrm{InforNCE}_{RGB}$ $\rightarrow$  $\textrm{MSF}_{RGB}$  & RGB & 400+200 & 39.6 & 50.8 \\
        
        \midrule
        \rowcolor{LightYellow}
        $\textrm{InforNCE}_{Flow}$*  & Flow & 400 & 45.3 & 66.1 \\
        \rowcolor{LightYellow}
        $\textrm{InforNCE}_{Flow}$ $\rightarrow$  $\textrm{MSF}_{Flow}$  & Flow & 400+200 & 47.3 & 64.7 \\
        
        \midrule
        \rowcolor{LightYellow}
        $\textrm{InforNCE}_{Flow}$*  & Flow & 400 & 45.3 & 66.1 \\
        \rowcolor{LightCyan}
        $\textrm{InforNCE}_{Flow}$ $\rightarrow$ $\textrm{CoCLR}_{RGB}^{Flow}$ & RGB & 400+100 &  49.8 & 61.0 \\
        
        \rowcolor{LightYellow}
        $\textrm{InforNCE}_{Flow}$ $\rightarrow$ $\textrm{CoCLR}_{RGB}^{Flow}$ $\rightarrow$ $\textrm{CoCLR}_{Flow}^{RGB}$ & Flow & 400+100+100 & 50.0 & 67.3 \\
        \rowcolor{LightCyan}
        \midrule
        \rowcolor{LightYellow}
        $\textrm{InforNCE}_{Flow}$*  & Flow & 400 & 45.3 & 66.1 \\
        \rowcolor{LightCyan}
        $\textrm{InforNCE}_{Flow}$ $\rightarrow$ $\textrm{CMSF}_{RGB}^{Flow}$ & RGB & 400+100 & 45.8 & 58.1\\
        \rowcolor{LightYellow}
        $\textrm{InforNCE}_{Flow}$ $\rightarrow$ $\textrm{CMSF}_{RGB}^{Flow}$ $\rightarrow$ $\textrm{CMSF}_{Flow}^{RGB}$ & Flow & 400+100+100 & 54.1 & 71.2\\
        \bottomrule
    \end{tabular}
    }
    \vspace{-.1in}
\end{table}

\section{Results on Semi-supervised Setting}
\label{results_semisup}
Unless specified, we use the ImageNet100 dataset for all the ablations on the semi-supervised setting for faster experimentation. 

\subsection{Role of Confidence Threshold in Pseudo-labeling} 
\begin{figure}
    \centering
    \includegraphics[width=0.75\linewidth]{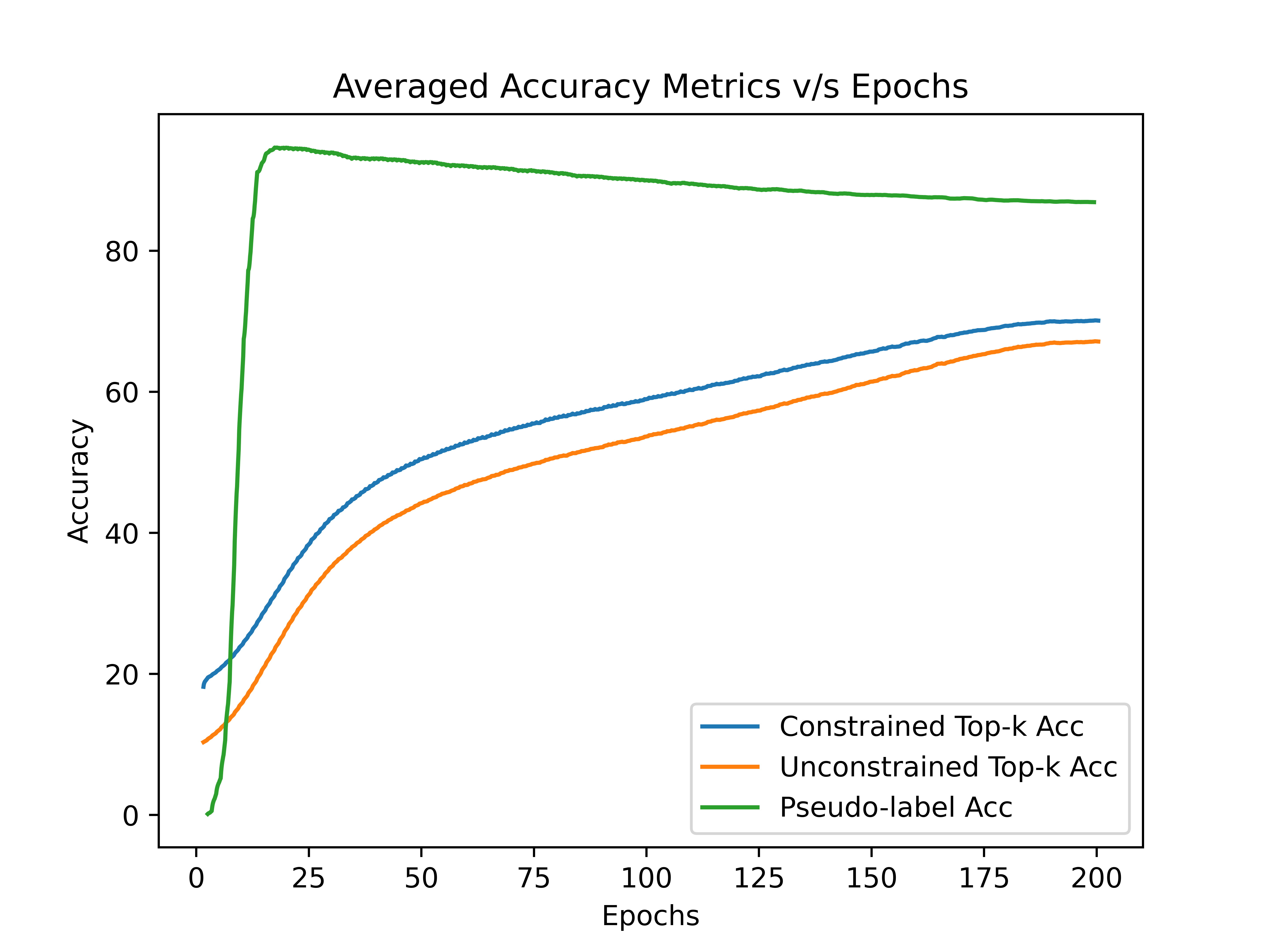}
    \caption{\textbf{Unconstrained NN accuracy in semi-supervised:} For analysis, we track the pseudo-labeling accuracy and accuracy of top-k neighbors chosen with and w/o applying the pseudo-label based constraint during training. The more accurate constrained NNs provide a better training signal. Pseudo-label accuracy on confident samples remains high throughout training, decreasing slightly as more confident samples are added.}
    \label{fig:semi_sup_purity}
\end{figure}

\begin{figure*}[ht!]
    \centering
    \includegraphics[width=1.0\linewidth]{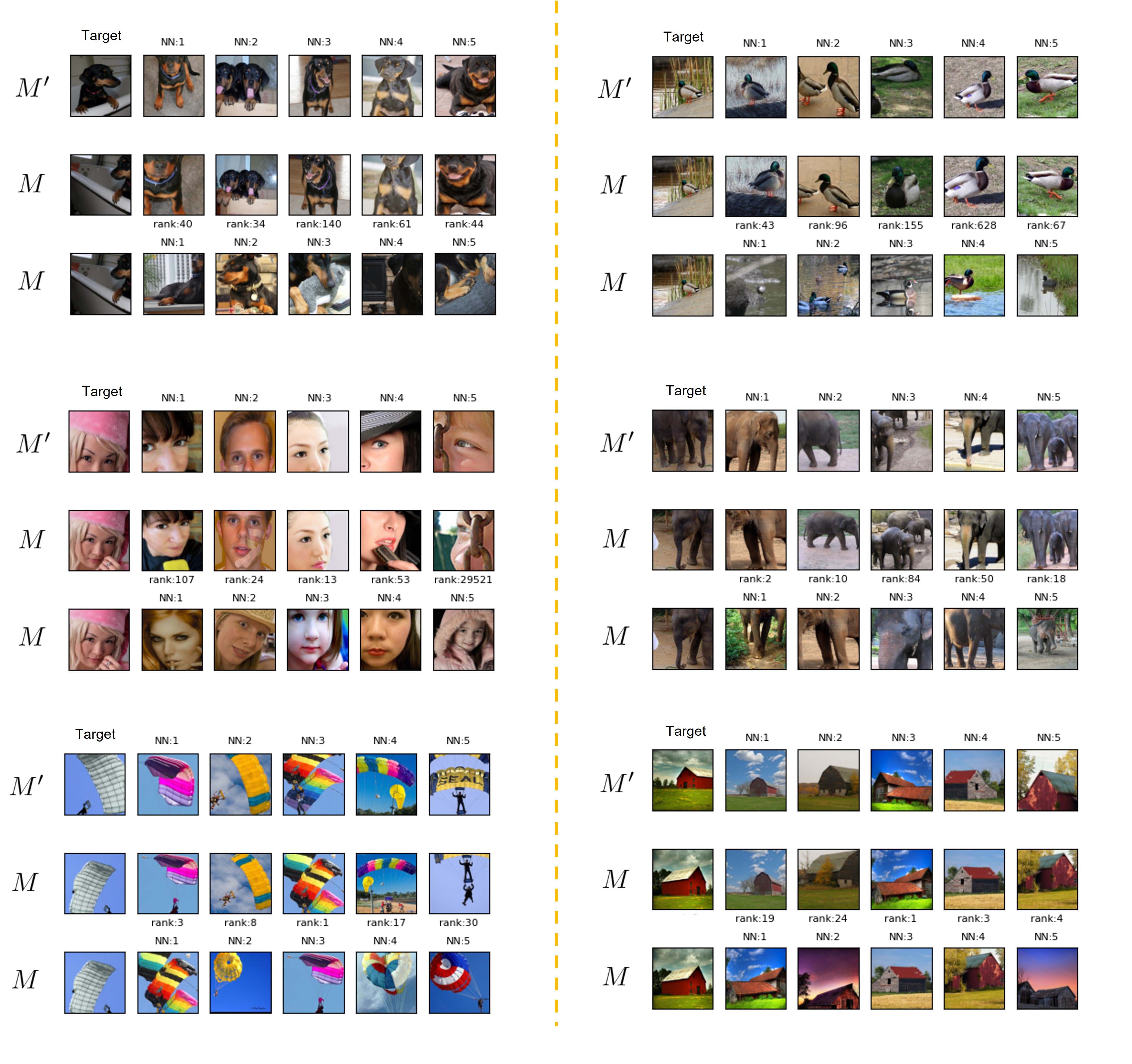}
    \caption{{\bf $\textrm{CMSF}_{\textrm{self}}$ nearest neighbor selection:} We use epoch 100 of $\textrm{CMSF}_{\textrm{self}}$ to visualize Top-$5$ NN from primary ($M$) and auxiliary ($M'$) memory banks. $M$ stores features for the current epoch while $M'$ contains representations from a different augmentation of the same image instance from the previous epoch. First row shows the target image and its top-$5$ NNs from the auxiliary memory bank $M'$. Samples of the second row are the images in $M$ corresponding to the ones in row 1. Thus, rows 1 and 2 contain different augmentations of the same image instances. We also report their rank in $M$ in row 2. The last row contains the top-$5$ NNs in $M$. Note that constrained samples in $M$ (second row), have high rank while they are semantically similar to the target.}
    \label{fig:vis_nn_2q}
\end{figure*}

We use a MLP classification head to predict pseudo-labels for the unlabeled set. As shown in Fig.~\ref{fig:semi_sup_purity}, the accuracy of the classifier is low in the initial stages and improves as training progresses. Using constraints from incorrectly labeled samples might affect the learning process. Thus, we use confidence (class probabilities) based thresholding to select the samples to be used for pseudo-labeling. Only those samples with confidence higher than the threshold are assigned a pseudo-label. Fig.~\ref{fig:semi_sup_purity} shows that the accuracy of the classifier on the confident samples remains high throughout training, limiting the number of incorrect pseudo-labels. Results for threshold value ($t$) selection are shown in table~\ref{tab:conf_th}. As expected, $t=0$ (i.e, no thresholding) performs poorly compared to higher threshold values. Pseudo-labeling accuracy increases with increasing value of $t$ and the best result is observed for $t=0.9$. While further increase in $t$ could result in higher pseudo-labeling accuracy, it would also mean that fewer samples are assigned pseudo-labels. Thus, we use $t=0.9$ in all our experiments on ImageNet100. Since ImageNet-1k has ten times more classes, we reduce the value to $0.85$ for all our experiments on ImageNet-1k. \\

\begin{table}[]
    \centering
    \caption{\textbf{Role of confidence threshold in pseudo-labeling (ImageNet100 results): }Using confidence based threshold to pseudo-label helps improve performance by eliminating noisy pseudo-labels. A higher threshold value results in higher pseudo-label accuracy but also limits the number of samples that participate in constraint selection. We set the value of $t$ to $0.9$ on the ImageNet100 dataset and to $0.85$ on the more diverse (1000 classes) ImageNet-1k dataset.}
    \begin{tabular}{cccc}
        \toprule
        Threshold ($t$) & 1-NN & 20-NN & Top-1 \\
        \midrule
        0   & 67.9 & 71.2 & 76.2 \\
        0.7 & 67.9 & 72.3 & 77.1 \\
        0.9 & \textbf{69.0} & \textbf{72.7} & \textbf{77.5} \\
        \bottomrule
    \end{tabular}
    \label{tab:conf_th}
\end{table}

\subsection{Effect of Caching on Pseudo-label Training}

In addition to optimizing the query encoder network using CMSF loss, we train the pseudo-label classifier head at the end of each epoch of query encoder training. Each round of pseudo-label training entails $40$ epochs of classifier head training on the supervised subset of the data ($10\%$). While the time required for backward pass is minimal since only the MLP head is updated, forward pass through the encoder adds significant computational overhead. We thus employ encoder feature caching to overcome this issue. We experiment with two caching settings - offline caching and online caching. In offline caching, encoder features for all the supervised samples are calculated once at the beginning of pseudo-label training and kept fixed for the remaining $39$ epochs. In online caching, encoder features for the supervised samples are cached for each mini-batch during the query network training. Similar to offline caching, these features are then fixed and used throughout the 40 epochs of pseudo-labeling network training. 
Offline technique requires one epoch of forward pass through the encoder, but has the advantage of using the most recent model parameters for feature calculation. Online caching results in features for different images being calculated using different encoder parameters. We observe that both these settings perform similarly on the ImageNet100 dataset (refer table~\ref{tab:pseudo_cache_schemes}). We thus use the online version in all our experiments since it has almost no overhead. With this setting, pseudo-label training increases the training time of each epoch approximately by just 40 seconds. \\

\begin{table}[]
    \centering
    \caption{\textbf{Feature caching for pseudo-label classifier training (ImageNet100 results): }We experiment two different caching schemes for pseudo-label training - offline and online. In offline caching, the features are calculated once at the beginning each round of pseudo-label training while in the online setting, the features are cached for each mini-batch during query encoder training. Since both approaches have similar performance, we use the online version since it has minimal computational overhead.}
    \begin{tabular}{lccc}
        \toprule
        Method & 1-NN & 20-NN & Top-1 \\
        \midrule
        Offline Caching & 67.9 & 71.2 & 76.2 \\
        Online Caching  & 66.5 & 71.9 & 76.0 \\
        \bottomrule   
    \end{tabular}
    \label{tab:pseudo_cache_schemes}
\end{table}

\subsection{Pseudo-label Classifier Selection}

The classifier used to generate pseudo-labels plays a crucial role in obtaining effective constraint sets for CMSF$_{\textrm{semi}}$. We experiment with two classification techniques - $k$-NN classifier and MLP classifier trained with cross-entropy loss. Results on ImageNet100 dataset are shown in table~\ref{tab:mlp_arch}. $k$-NN classifier has lower pseudo-labeling accuracy and thus results in poorer performance. We additionally experiment with linear, two and three layer architectures for the MLP classifier head. As shown in table~\ref{tab:mlp_arch}, multi-layer head significantly outperform the linear classifier. Since there is minimal difference in performance of two and three layer MLPs, we use a two layer MLP head in all our experiments. \\

\begin{table}[]
    \centering
    \caption{\textbf{Pseudo-label classifier selection (ImageNet100 results): }We experiment with different classifier methods and architectures for pseudo-label prediction. Linear layer or multi-layer perceptron (MLP) heads trained using cross-entropy loss on the supervised examples outperform a $k$-NN classifier. MLP classifiers achieve higher accuracy on the pseudo-labeling task on both the train and test sets. We use a two layer MLP head based classifier in all our experiments. }
    \begin{tabular}{lccc}
        \toprule
        Pseudo-label Classifier & 1-NN & 20-NN & Top-1 \\
        \midrule
        k-NN Classifier    & 64.9 & 69.5 & 74.7 \\
        Linear Classifier  & 65.6 & 70.0 & 75.5 \\
        2 Layer MLP Head   & 67.9 & 71.2 & 76.2 \\
        3 Layer MLP Head   & 67.1 & 71.0 & 76.1 \\
        \bottomrule
    \end{tabular}
    \label{tab:mlp_arch}
\end{table}

\subsection{Fine-tuning without Pseudo-labels}

Since we do not explicitly optimizer our encoder networks on the label classification task in the pre-training stage, we perform two-stage fine-tuning. We initially fine-tune the pretrained model on only the supervised samples and use the fine-tuned model to obtain pseudo-labels for the unsupervised ones. The combined data is then used to fine-tune the network again. In table~\ref{tab:fine_tune}, we present results with just a single round of fine-tuning with the $10\%$ supervised samples on ImageNet-1k. Two rounds of fine-tuning provides a small improvement in performance over the single-stage version. \\

\begin{table}[]
    \centering
    \caption{\textbf{Role of network fine-tuning on classification performance (ImageNet-1k results): }We evaluate the trained models using the linear evaluation technique commonly employed for evaluating self-supervised approaches and entire network fine-tuning as performed in semi-supervised methods. Both the methods use $10\%$ of the dataset as supervision. We observe an increase in classification performance when both the encoder and MLP classifier are fine-tuned. }
    \begin{tabular}{lc}
        \toprule
        Fine-tune Method & Top-1 \\
        \midrule
        Linear layer training  & 76.5 \\
        Full network fine-tune & 76.9 \\
        \bottomrule
    \end{tabular}
    \label{tab:fine_tune}
\end{table}

\section{Results on Supervised Setting}
\label{results_sup}

\subsection{Coarse-grained ImageNet}
\label{sec:coarse_imagenet}
$\textrm{CMSF}_{\textrm{sup}}$ top-$k$ groups together only top-$k$ neighbors and thus can help in preserving the latent structure of the data compared to top-$all$. To verify this, we consider a dataset with coarse-grained labels where this difference is pronounced. ImageNet dataset was constructed using the WordNet hierarchy. Consider the subtree of WordNet that contains all the $1000$ categories from ImageNet-1k as the leaf nodes. To obtain a coarse-grained version, we merge each category in the leaf node to its parent node. 
After merging, we further ensure that no two of the newly obtained super-classes are in the same path in the graph by merging the descendant into the ancestor class. The total number of classes is thus reduced from $1000$ in ImageNet-1k to $93$ in our ImageNet-coarse. We train CMSF$_{\textrm{self}}$and the baseline approaches in a supervised manner using the coarse labels and then evaluate on the fine-grained (i.e., original) labels on ImageNet-1k validation set. The training settings remain same as that of CMSF$_{\textrm{sup}}$ in Sec.3.2 of main submission.

In Table~\ref{tab:coarse_imagenet_suppl} we compare the top-$all$, top-$1000$ and top-$k$ variants on the coarse grained version of ImageNet. We consider the top-$1000$ variant to limit the effect of dataset imbalance introduced due to the merging of classes. $\textrm{CMSF}_{\textrm{sup}}$ top-$k$ sees a minor drop in performance compared to training on ImageNet-1k. However, methods in which most or all samples in a class are explicitly brought closer - $\textrm{CMSF}_{\textrm{sup}}$ top-$all$ and top-$1000$, cross-entropy and supervised contrastive - see a huge drop in accuracy.  

\begin{table}[t]
    \centering
    \caption{\textbf{Supervised learning on coarse grained ImageNet:} We train on the coarse grained version of ImageNet (93 super categories) and perform linear evaluation on the original ImageNet-1k validation set with fine-grained labels (1000 categories). $\textrm{CMSF}_{\textrm{sup}}$ top-$10$ outperforms all other variants and baselines.}
    \scalebox{1.0}{
    \begin{tabular}{lccccc}
         \toprule
         \multirow{2}{*}{Train Dataset}                    & \multicolumn{5}{c}{ImageNet-1k Validation Set} \\
                & Xent & SupCon & $\textrm{CMSF}_{\textrm{sup}}$ & $\textrm{CMSF}_{\textrm{sup}}$ & $\textrm{CMSF}_{\textrm{sup}}$ \\
                & & & top-$all$ & top-$1000$ & top-$10$ \\
         \midrule
         ImageNet-1k         & 77.2 & \textbf{77.5}   & 75.7    & -       & 76.4 \\
         ImageNet-coarse     & 61.4 & 58.7   & 67.0   & 71.0    & \textbf{74.2} \\
    \bottomrule
    \end{tabular}}
    \label{tab:coarse_imagenet_suppl}
\end{table}

\subsection{Ablations}
\label{sec:ablations}

We explore different design choices and parameters of our method and baselines. We add the techniques used for our methods to the baselines to isolate the effect of different losses. The results are reported in Table \ref{tab:all_ablations}. Training and evaluation details are the same as in Section 3.2 of the main submission.

\begin{table}[]
    \begin{center}

    \caption{{\bf Ablations of baselines and CMSF$_{\textrm{sup}}$:} All experiments use 200 epochs if not mentioned and use ImageNet-1k dataset. \textbf{(a)} More epochs does not improve transfer accuracy for Xent. Thus, the model available from PyTorch \cite{official_pytorch_models} (last row) has the best transfer accuracy; \textbf{(b)} We add components of our method to improve SupCon baseline. The baseline implementation of SupCon uses std. aug and 16k memory size and it does not include the target embedding $u$ in the positive set. \textbf{(c)} We find that our method is not very sensitive to the size of memory bank or top-$k$ in supervised settings; \textbf{(d)} Interestingly, excluding the target embedding $u$ from $C$ does not hurts the results. Note that when we do not include the target, the nearest neighbors are still chosen based on the distance to the target, so they will be close to the target. }
    \begin{tabular}{lcc}
        \toprule
        Method & Mean & Linear \\
        & Trans & IN-1k \\
        
        \midrule
        \textit{(a) Xent} & & \\
        lr=$0.05$, cos, epochs=$200$, strong aug. & 71.5 & 77.2 \\
        lr=$0.05$, cos, epochs=$200$, std. aug. & 71.0 & 77.3 \\
        lr=$0.10$, cos, epochs=$200$, strong aug. & 72.3 & 77.1 \\
        lr=$0.05$, cos, epochs=$90$, std. aug. & 72.4 & 76.8 \\
        lr=$0.10$, cos, epochs=$90$, std. aug. & 74.0 & 76.7 \\
        lr=$0.10$, step, epochs=$90$, std. aug. & 74.9 & 76.2 \\
        
        \midrule
        \textit{(b) SupCon} & & \\
        Base SupCon & 77.2 & 77.9 \\
        + change to strong aug. & 77.9 & 77.4 \\
        + add target to positive set & 77.8 & 77.4 \\
        + change to weak/strong aug. & 77.8 & 77.2 \\
        + increase mem size to 128k & 78.4 & 77.5 \\

        \midrule
        \textit{(c) CMSF$_{\textrm{sup}}$} & & \\
        top-$1$ \small{(BYOL-asym)}  & 74.3 & 69.3 \\
        mem=128k, top-$2$ & 78.4 & 76.2 \\
        mem=128k, top-$10$ & 80.1 & 76.4 \\
        mem=128k, top-$20$ & 79.9 & 76.3 \\
        mem=128k, top-$all$ & 80.1 & 75.7 \\
        mem=512k, top-$10$ & 79.9 & 76.2 \\
        mem=512k, top-$20$ & 80.1 & 76.3 \\
        
        \midrule
        \textit{(d) CMSF$_{\textrm{sup}}$} & & \\
        target in top-$10$ & 80.1 & 76.4 \\
        target not in top-$10$ & 80.3 & 76.4 \\
        
        \bottomrule
    \end{tabular}

    \label{tab:all_ablations}
    \end{center}
\end{table}

\setlength{\tabcolsep}{2pt}
\begin{table*}[ht!]
    \centering
    \caption{\textbf{ResNet50 backbone training FLOPs calculation:}We provide the number of forward and backward passes per image (including multi-crops) and the total such passes for the entire training stage. Mean Shift and the proposed constrained mean shift methods have the least compute requirement among all approaches. Eq.~\ref{eqn:comp_calc} provides the formula to calculate the total number of passes and FLOPs. In PAWS, \textit{sup} refers to the size of the support set in the mini-batch.}
    \vspace{-0.10in}
    \scalebox{0.93}{
    \begin{tabular}{lccc|ccc|ccccc}
        \toprule
        Method & \multicolumn{3}{c}{Unlabeled} & \multicolumn{3}{c}{Labeled} & Mini- & Iters & Epochs & Total & FLOPs \\
         & Fwd & Bwd & BS & Fwd & Bwd & BS & Batch & per & & Pass &  \\
         &     &     &    &     &     &    &       &      epoch   & & ($\times10^{8}$) & ($\times10^{18}$)\\
        \midrule
        Mean Shift~\cite{koohpayegani2021mean}          & 2   & 1   & 256  &   &   &      & 768   & 5004 & 200  & 7.7  & 4     \\       
        BYOL~\cite{grill2020bootstrap}                  & 4   & 2   & 4096 &   &   &      & 24576 & 312  & 1000 & 76.7 & 40     \\       
        SwAV~\cite{caron2020unsupervised}               & 3.1 & 3.1 & 4096 &   &   &      & 25395 & 312  & 800  & 63.4 & 37    \\       
        SimCLRv2~\cite{chen2020big}                     & 2   & 2   & 4096 &   &   &      & 16384 & 312  & 800  & 40.9 & 16     \\        
        UDA$^{\dagger}$~\cite{xie2019unsupervised}      & 2   & 1   & 15360& 1 & 1 & 512  & 47104 & \multicolumn{2}{l}{40000}    & 18.8 & 10      \\  
        FixMatch$^{\dagger}$~\cite{sohn2020fixmatch}    & 2   & 1   & 5120 & 1 & 1 & 1024 & 17408 & 250  & 300  & 13.1 & 70       \\  
        MPL$^{\dagger}$~\cite{pham2021meta}             & 3   & 2   & 2048 & 2 & 2 & 128  & 10752 & \multicolumn{2}{l}{500000}   & 53.8 & 30     \\  
        PAWS (sup=6720)~\cite{assran2021semi}       & 3.1 & 3.1 & 4096 & 1 & 1 & 6720 & 38835 & 312  & 300  & 36.6 & 21  \\     
        PAWS (sup=1680)~\cite{assran2021semi}       & 3.1 & 3.1 & 256  & 1 & 1 & 1680 & 4947  & 5004 & 100  & 24.8 & 15  \\       
        PAWS (sup=400)~\cite{assran2021semi}        & 3.1 & 3.1 & 256  & 1 & 1 & 400  & 2387  & 5004 & 100  & 12.0 & 7   \\       
        \rowcolor{LightYellow}                                                                                              
        CMSF$_{\textrm{semi}}$-basic                                   & 2   & 1   & 256  &   &   &      & 768   & 5004 & 200  & 7.7  & 4    \\         
        \rowcolor{LightYellow}                                                                                              
        CMSF$_{\textrm{semi}}$                                     & 2   & 1   & 256  &   &   &      & 768   & 5004 & 200  & 7.7  & 4    \\   
        \rowcolor{LightYellow}                                                                                              
        CMSF$_{\textrm{semi}}$-mix prec.                       & 2   & 1   & 768  &   &   &      & 2304  & 1668 & 200  & 7.7  & 4    \\   
        \bottomrule
    \end{tabular}
    }
    \label{tab:flops_calc_semi_sup}
\end{table*}

\section{Implementation Details}
\label{impl_details}

\subsection{Transfer Learning}

We use the LBFGS optimizer (max\_iter=20, and history\_size=10) along with the Optuna library \cite{akiba2019optuna} in the Ray hyperparameter tuning framework \cite{liaw2018tune}. Each dataset gets a budget of 200 trials to pick the best parameters on validation set. The final accuracy is reported on a held-out test set by training the model on the train+val split using the best hyperparameters. The hyperparameters and their search spaces (in loguniform) are as follows: iterations $\in [0, 10^3]$, lr $\in [10^{-6}, 1]$, and weight decay $\in [10^{-9}, 1]$. We also show that we can reproduce the transfer results for BYOL \cite{grill2020bootstrap} and SimCLR \cite{chen2020simple} with our framework. The features are extracted with the following pre-processing for all datasets: resize shorter side to 256, take a center crop of size 224, and normalize with ImageNet statistics. No training time augmentation was used.

\begin{table*}[]
    \begin{center}
    
    \caption{\textbf{Noisy supervised setting on ImageNet-100:} Our method is more robust to noisy annotation compared to Xent and SupCon. The top-$all$ variant suffers greater degradation compared to top-$10$ since all images from a single category are not guaranteed to be semantically related in the noisy setting.\\}
    \vspace{-0.25in}
    \scalebox{0.83}{
    \begin{tabular}{l|r|c|c|c|c|c|c|c|c|c|c||c|c}
    \toprule
    Method & Noise & Food & CIFAR & CIFAR & SUN & Cars & Air- & DTD & Pets & Calt. & Flwr & \textbf{Mean} & \textbf{Linear} \\
    & & 101 & 10 & 100 & 397 & 196 & craft &  &  & 101 & 102 & \textbf{Trans} & \textbf{IN-100} \\
    \midrule
    Xent & 0\% & 53.6 & 81.9 & 61.1 & 37.8 & 25.7 & 29.5 & 56.9 & 69.7 & 70.2 & 82.3 & 56.9 & 85.7 \\
    SupCon & 0\% & 61.5 & 88.7 & 69.0 & 49.1 & 51.6 & 48.2 & 65.4 & 81.0 & 87.0 & 89.8 & 69.1 & 86.9 \\
    CMSF$_{\textrm{sup}}$ \small{top-$all$} & 0\% & 61.6 & 88.2 & 68.5 & 49.9 & 54.6 & 52.7 & 64.7 & 82.2 & 89.6 & 89.1 & 70.1 & 84.9 \\
    CMSF$_{\textrm{sup}}$ \small{top-$10$} & 0\% & 62.6 & 86.8 & 66.2 & 50.5 & 54.7 & 51.0 & 64.6 & 82.4 & 88.5 & 90.4 & 69.8 & 85.0 \\
    \midrule
    Xent & 5\% & 46.5 & 81.1 & 58.1 & 35.8 & 27.5 & 36.0 & 58.7 & 67.5 & 73.3 & 77.0 & 56.1 & 81.5 \\
    SupCon & 5\% & 60.0 & 87.1 & 66.4 & 48.2 & 52.1 & 47.8 & 65.1 & 80.8 & 85.7 & 89.3 & 68.3 & 85.7 \\
    CMSF$_{\textrm{sup}}$ \small{top-$all$} & 5\% &  60.3 & 87.5 & 66.4 & 49.1 & 55.5 & 53.0 & 64.8 & 80.9 & 87.3 & 89.9 & 69.5 & 84.4 \\
    CMSF$_{\textrm{sup}}$ \small{top-$10$} & 5\% & 61.6 & 86.8 & 67.4 & 49.6 & 55.8 & 51.2 & 63.4 & 81.5 & 86.7 & 90.6 & 69.5 & 84.7 \\
    \midrule
    Xent & 10\% & 44.1 & 79.5 & 56.1 & 32.4 & 26.1 & 34.5 & 56.1 & 69.7 & 72.5 & 75.1 & 54.6 & 79.6 \\
    SupCon & 10\% & 58.8 & 85.8 & 66.4 & 47.0 & 50.6 & 47.7 & 65.3 & 79.8 & 85.0 & 89.1 & 67.6 & 84.0 \\
    CMSF$_{\textrm{sup}}$ \small{top-$all$} & 10\% & 59.4 & 86.4 & 66.0 & 48.8 & 55.0 & 51.4 & 64.7 & 80.1 & 87.8 & 89.0 & 68.9 & 83.1 \\
    CMSF$_{\textrm{sup}}$ \small{top-$10$} & 10\% & 60.9 & 87.2 & 66.9 & 49.4 & 54.2 & 51.4 & 65.5 & 80.6 & 88.5 & 90.0 & 69.5 & 83.8 \\
    \midrule
    Xent & 25\% & 49.0 & 77.2 & 54.5 & 30.6 & 25.9 & 30.7 & 53.1 & 66.6 & 64.1 & 77.8 & 53.0 & 75.2\\
    SupCon & 25\% & 55.6 & 84.9 & 63.4 & 43.1 & 43.9 & 43.7 & 62.9 & 74.3 & 82.1 & 86.8 & 64.1 & 81.1 \\
    CMSF$_{\textrm{sup}}$ \small{top-$all$} & 25\% & 56.4 & 85.7 & 64.2 & 46.0 & 53.6 & 49.6 & 62.7 & 74.2 & 85.2 & 87.4 & 66.5 & 78.8 \\
    CMSF$_{\textrm{sup}}$ \small{top-$10$} & 25\% & 58.9 & 85.2 & 64.9 & 47.8 & 55.0 & 50.6 & 64.0 & 80.0 & 86.3 & 89.7 & 68.2 & 81.8 \\
    \midrule
    Xent & 50\% & 44.4 & 72.3 & 51.3 & 31.1 & 21.4 & 24.9 & 46.0 & 57.4 & 56.0 & 73.0 & 47.8 & 67.8 \\
    SupCon & 50\% & 30.8 & 64.9 & 38.9 & 24.2 & 13.6 & 20.5 & 45.5 & 55.2 & 60.1 & 59.2 & 41.3 & 69.0 \\
    CMSF$_{\textrm{sup}}$ \small{top-$all$} & 50\% & 44.7 & 79.3 & 54.9 & 35.2 & 35.7 & 41.2 & 54.9 & 54.6 & 75.3 & 75.1 & 55.1 & 61.6 \\
    CMSF$_{\textrm{sup}}$ \small{top-$10$} & 50\% & 58.7 & 85.7 & 64.2 & 47.5 & 51.6 & 50.5 & 62.0 & 77.3 & 86.8 & 70.1 & 65.4 & 80.1 \\
    \bottomrule
    \end{tabular}
    }
    \label{tab:main_noisy}
    \end{center}
\end{table*}

\begin{table}[ht!]
    \begin{center}
    
    \caption{\textbf{Transfer dataset details:} Train, val, and test splits of the transfer datasets are listed in this table. \textbf{Test split: } We follow the details in \cite{koohpayegani2021mean}. For Aircraft, DTD, and Flowers datasets, we use the provided test sets. For Sun397, Cars, CIFAR-10, CIFAR-100, Food101, and Pets datasets, we use the provided val set as the hold-out test set. For Caltech-101, 30 random images per category are used as the hold-out test set. \textbf{Val split: } For DTD and Flowers, we use the provided val sets. For other datasets, the val set is randomly sampled from the train set. For transfer setup, to be close to BYOL \cite{grill2020bootstrap}, the following val set splitting strategies have been used for each dataset: Aircraft: 20\% samples per class. Caltech-101: 5 samples per class. Cars: 20\% samples per class. CIFAR-100: 50 samples per class. CIFAR-10: 50 samples per class. Food101: 75 samples per class. Pets: 20 samples per class. Sun397: 10 samples per class. \textbf{Accuracy measure: }\textit{Top-1} refers to top-1 accuracy while \textit{Mean} refers to mean per-class accuracy.\\}
    \scalebox{1.00}{
    \begin{tabular}{lccccccc}
        \toprule
        Dataset & Classes & Train & Val       & Test    & Accuracy & Test set \\
                &         & samples & samples & samples & measure  & provided \\
        \midrule
        Food101 \cite{food101}         & 101 & 68175 & 7575 & 25250 & Top-1 & - \\
        CIFAR-10 \cite{cifar}          & 10  & 49500 & 500  & 10000 & Top-1 & - \\
        CIFAR-100 \cite{cifar}         & 100 & 45000 & 5000 & 10000 & Top-1 & -\\
        Sun397 (split 1) \cite{sun397} & 397 & 15880 & 3970 & 19850 & Top-1 & - \\
        Cars \cite{carsdataset}        & 196 & 6509  & 1635 & 8041  & Top-1 & - \\
        Aircraft \cite{aircraft}       & 100 & 5367  & 1300 & 3333  & Mean  & Yes \\
        DTD (split 1) \cite{dtd}       & 47  & 1880  & 1880 & 1880  & Top-1 & Yes \\
        Pets \cite{pets}               & 37  & 2940  & 740  & 3669  & Mean  & - \\
        Caltech-101 \cite{caltech101}  & 101 & 2550  & 510  & 6084  & Mean  & - \\
        Flowers \cite{flowers}         & 102 & 1020  & 1020 & 6149  & Mean  & Yes \\
        \bottomrule
    \end{tabular}
    }
    \label{tab:appendix_transfer_dset_details}
    \end{center}
\end{table}

\subsection{Supervised Setting}
\subsubsection{Implementation Details of Baselines}

\noindent \textbf{SupCon: }The MLP architecture for SupCon baseline is: linear (2048x2048), batch norm, ReLU, and linear (2048x128). To optimize the SupCon baseline, following \cite{khosla2020supervised}, we use the first 10 epochs for learning-rate warmup. For both SupCon and ProtoNW, the temperature is $0.1$. 

\noindent \textbf{Prototypical Networks (ProtoNW): } In order to further study the effect of contrast, we design another contrastive version of our top-$all$ variation. We calculate a prototype for each class by averaging all its instances in the memory bank. Then, similar to prototypical networks \cite{snell2017prototypical}, we compare the input with all prototypes by passing their temperature-scaled cosine distance through a SoftMax layer to get probabilities. Finally, we minimize the cross-entropy loss. Note that this method is still contrastive in nature because of the SoftMax operation.

\subsection{Semi-supervised Setting}

\noindent \textbf{Pretraining:} Similar to the self-supervised setting, we train the network for 200 epochs using SGD optimizer (batch size=256, lr=0.05, momentum=0.9, weight decay=1e-4). Ten nearest neighbors are chosen from the constraint set for loss calculation. The size of memory bank is set to 128000. We train the pseudo-label classifier using an additional SGD optimizer (batch size=256, lr=0.01, momentum=0.9, weight decay=1e-4) for 10 epochs at the end of each epoch of query encoder training. A confidence threshold value of $0.85$ is used to assign pseudo-labels to the unlabeled samples. \\

\noindent \textbf{Fine-tuning:} In addition to pretraining, we use a two layer MLP atop the CNN backbone and fine-tune the entire network on the supervised subset for 20 epochs. This fine-tuned network is used to pseudo-label the unlabeled set with a confidence threshold of $0.9$. Samples above the threshold are combined with the supervised set for a second round of fine-tuning for 20 epochs. We observe that nearly one third of the samples in the dataset have confidence higher than the threshold at the end of the first fine-tuning stage. We use a SGD optimizer (batch size=256, lr=0.005, momentum=0.9, weight decay=1e-4) for both the fine-tuning stages. The learning rate is multiplied by $0.1$ at the end of epoch 15. \\

\noindent \textbf{Calculation of forward and backward FLOPs:} In figure 1 of the main submission, we present a plot of top-1 accuracy against total compute and resources for various semi-supervised approaches. Here (table~\ref{tab:flops_calc_semi_sup}) we present the calculation of the forward and backward FLOPS for each of the methods. We set the backward FLOPs to be twice the forward number of FLOPs~\cite{he2016deep} for a single image and the total FLOPs to be the sum of forward and backward pass FLOPs for the entire training. We use a value of 3.9 GFLOPs for a single forward pass of $224\times 224$ resolution image through the ResNet50 backbone~\cite{hu2018squeeze}. Additional compute due to the use of multi-crops are accounted for. A scalar multiplier of ${(\frac{K}{224})}^2$ is used for images of resolution $K\times K$ (e.g., using one ($96\times96$) image would be equivalent to $0.184$ image of resolution ($224\times224$)). However, we do not consider the floating point precision (mixed or full precision) in our calculations. We show that similar performance can be achieved by using both automatic mixed precision and full precision floating point during training (table 4, main submission) and thus focus the compute calculation on the total number of forward and backward passes. Eq.~\ref{eqn:comp_calc} provides the formula to calculate the total number of training passes and FLOPs.

\begin{equation}
\begin{split}
   \textrm{Fwd mini-batch}  &= (\textrm{Unlabeled fwd crops} * \textrm{Unlabeled batch-size})\\ &+ (\textrm{Labeled fwd crops} * \textrm{Labeled batch-size}) \\
   \textrm{Bwd mini-batch} &= (\textrm{Unlabeled bwd crops} * \textrm{Unlabeled batch-size})\\ &+ (\textrm{Labeled bwd crops} * \textrm{Labeled batch-size}) \\
   \textrm{Fwd passes}           &= \textrm{Fwd mini-batch} * \textrm{Iterations per epoch} * \textrm{Epochs} \\
   \textrm{Bwd passes}          &= \textrm{Bwd mini-batch} * \textrm{Iterations per epoch} * \textrm{Epochs} \\
   \textrm{Total FLOPs}              &= (\textrm{Fwd passes} + 2 * \textrm{Bwd passes}) * (3.9\times10^{9})
\end{split}
\label{eqn:comp_calc}
\end{equation}

\clearpage
\bibliographystyle{splncs04}
\bibliography{egbib}